%% file: main.tex
\documentclass{article} % For LaTeX2e

\usepackage[accepted]{icml2018}

\icmltitlerunning{Deep Variational Reinforcement Learning}

\input{include/packages}

\input{include/colors}
\input{include/todo}
\author{Max, Luisa, TuanAnh, Frank, Shimon}

\begin{document}

\twocolumn[
\icmltitle{Deep Variational Reinforcement Learning for POMDPs}

% \maketitle

% It is OKAY to include author information, even for blind
% submissions: the style file will automatically remove it for you
% unless you've provided the [accepted] option to the icml2018
% package.

% List of affiliations: The first argument should be a (short)
% identifier you will use later to specify author affiliations
% Academic affiliations should list Department, University, City, Region, Country
% Industry affiliations should list Company, City, Region, Country

% You can specify symbols, otherwise they are numbered in order.
% Ideally, you should not use this facility. Affiliations will be numbered
% in order of appearance and this is the preferred way.
\icmlsetsymbol{equal}{*}

\begin{icmlauthorlist}
\icmlauthor{Maximilian Igl}{ox}
\icmlauthor{Luisa Zintgraf}{ox}
\icmlauthor{Tuan Anh Le}{ox}
\icmlauthor{Frank Wood}{ubc}
\icmlauthor{Shimon Whiteson}{ox}
\end{icmlauthorlist}

\icmlaffiliation{ox}{University of Oxford, United Kingdom}
\icmlaffiliation{ubc}{University of British Columbia, Canada}

\icmlcorrespondingauthor{Maximilian Igl}{maximilian.igl@eng.ox.ac.uk}

% You may provide any keywords that you
% find helpful for describing your paper; these are used to populate
% the "keywords" metadata in the PDF but will not be shown in the document
\icmlkeywords{Machine Learning, ICML, Reinforcement Learning, POMDP, Partially
  Observable Markov Decision Process, Particle Filter, Deep Learning}

\vskip 0.3in
]

% this must go after the closing bracket ] following \twocolumn[ ...

% This command actually creates the footnote in the first column
% listing the affiliations and the copyright notice.
% The command takes one argument, which is text to display at the start of the footnote.
% The \icmlEqualContribution command is standard text for equal contribution.
% Remove it (just {}) if you do not need this facility.

\printAffiliationsAndNotice{}  % leave blank if no need to mention equal contribution
% \printAffiliationsAndNotice{\icmlEqualContribution} % otherwise use the standard text.

\begin{abstract}
	Many real-world sequential decision making problems are partially observable by nature,
	and the environment model
  is typically unknown.
	Consequently, there is great need for reinforcement learning methods that can tackle such problems given only a stream of incomplete and noisy observations.
  In this paper, we propose \gls{DVRL}, which introduces an inductive bias that allows an agent to learn a
  generative model of the environment and perform inference in that model to effectively
  aggregate the available information.
  We develop an $n$-step approximation to the \gls{ELBO}, allowing the model to
  be trained jointly with the policy.
  This ensures that the latent state representation is suitable for the control task.
  In experiments on Mountain Hike and flickering Atari we show that our method
  outperforms previous approaches relying on recurrent neural networks to encode the past.
\end{abstract}

\input{core/1_intro}
\input{core/3_background}
\input{core/4_methods}
\input{core/2_related-work}

\input{core/5_experiments}

\input{core/7_conclusion}

\section*{Acknowledgements}
We would like to thank Wendelin Boehmer and Greg Farquar for useful discussions and feedback. The NVIDIA DGX-1
used for this research was donated by the NVIDIA corporation. M.\ Igl is supported
by the UK EPSRC CDT in Autonomous Intelligent Machines and Systems. L.\ Zintgraf is
supported by the Microsoft Research PhD Scholarship Program.
T. A. Le is supported by EPSRC DTA and Google (project code DF6700) studentships.
F.\ Wood is supported by DARPA PPAML through the U.S. AFRL under Cooperative Agreement FA8750-14-2-0006; Intel and DARPA D3M, under Cooperative
Agreement FA8750-17-2-0093.
S.\ Whiteson is supported by the European Research Council (ERC) under the European Union's Horizon 2020 research and innovation programme (grant agreement number 637713).

\clearpage
\newpage
\bibliography{bib/refs_max}
\bibliographystyle{icml2018}

\onecolumn
\appendix
\input{core/8_appendix_experiments}
\input{core/8_appendix_algorithms}

\end{document}

%% file: include/packages.tex
\usepackage{blindtext}

\usepackage{emptypage}

\usepackage{graphicx}

\usepackage{float}

\usepackage{amsmath}
\usepackage{amssymb}

\usepackage{gensymb}
\usepackage{textcomp}
\usepackage{setspace}

\usepackage{multirow}

\usepackage{natbib}

\usepackage[font=small,labelfont=bf]{caption}

\usepackage{etoolbox}

\usepackage{lipsum}

\input{include/colors}

\usepackage{booktabs}

\usepackage{tikz}
\usetikzlibrary{calc,trees,positioning,arrows,chains,shapes.geometric,%
  decorations.pathreplacing,decorations.pathmorphing,shapes,%
  matrix,shapes.symbols,fit,decorations,arrows.meta}

\PassOptionsToPackage{hyphens}{url}\usepackage{hyperref}
\hypersetup{
 unicode=false,          % non-Latin characters in Acrobat’s bookmarks
 pdftoolbar=true,        % show Acrobat’s toolbar?
 pdfmenubar=true,        % show Acrobat’s menu?
 pdffitwindow=false,     % window fit to page when opened
 pdfstartview={FitH},    % fits the width of the page to the window
 pdfauthor={Tim Rocktäschel},     % author
 pdfnewwindow=true,      % links in new window
 colorlinks=true,        % false: boxed links; true: colored links
 linkcolor=black,        % color of internal links (change box color with linkbordercolor)
 citecolor=black,    % color of links to bibliography
 filecolor=black,        % color of file links
 urlcolor=black          % color of external links
}

\usepackage{subcaption}

\usepackage{microtype}

\usepackage{hyperref}
\usepackage{url}

\usepackage{soul}

\usepackage{changepage}

\usepackage{xargs}                      % Use more than one optional parameter in a new commands
 
\usepackage{bm}

\usepackage[capitalize]{cleveref} % better autoref (\cref \Cref)
\crefformat{equation}{Eq.~#2#1#3}
\Crefformat{equation}{Equation~#2#1#3}
\crefrangeformat{equation}{Eqs.~#3#1#4 to~#5#2#6}
\Crefrangeformat{equation}{Equations~#3#1#4 to~#5#2#6}
\crefmultiformat{equation}{Eqs.~#2#1#3}%
{ and~#2#1#3}{, #2#1#3}{ and~#2#1#3}
\Crefmultiformat{equation}{Equations~#2#1#3}%
{ and~#2#1#3}{, #2#1#3}{ and~#2#1#3}

\usepackage{stmaryrd} % llbracket, rrbracket

\usepackage[colorinlistoftodos,prependcaption,textsize=small]{todonotes}
\usepackage{wrapfig}

\usepackage{afterpage}

\usepackage[normalem]{ulem}

\usepackage{microtype}
\usepackage{graphicx}
\usepackage{booktabs} % for professional tables

\usepackage{bbold}

\usepackage{nicefrac}

\usepackage[acronym,smallcaps,nowarn,section,nogroupskip,nonumberlist]{glossaries}
\glsdisablehyper{}
\newacronym{VAE}{vae}{variational autoencoder}
\newacronym{IWAE}{iwae}{importance weighted autoencoder}
\newacronym{IS}{is}{importance sampling}
\newacronym[firstplural=partially observable Markov decision processes]{POMDP}{pomdp}{partially observable Markov decision process}
\newacronym[firstplural=Markov decision processes]{MDP}{mdp}{Markov decision process}
\newacronym{RL}{rl}{reinforcement learning}
\newacronym{ADRQN}{adrqn}{action-specific deep recurrent Q-network}
\newacronym{(A)DRQN}{adrqn}{}
\newacronym{RNN}{rnn}{recurrent neural network}
\newacronym{VRNN}{vrnn}{variational recurrent neural network}
\newacronym{AESMC}{aesmc}{autoencoding sequential Monte Carlo}
\newacronym{SMC}{smc}{sequential Monte Carlo}
\newacronym{DVRL}{dvrl}{deep variational reinforcement learning}
\newacronym{A3C}{A3C}{asynchronous advantage actor-critic}
\newacronym{A2C}{A2C}{advantage actor-critic}
\glsunset{A2C}
\newacronym{ELBO}{elbo}{evidence lower bound}
\newacronym{DQN}{dqn}{deep Q-network}
\newacronym{DPFP}{dpfp}{deep particle filter based policy}
\newacronym{CNN}{cnn}{convolutional neural network}
\newacronym{DRQN}{drqn}{deep recurrent Q-network}
\newacronym{KL}{kl}{Kullback-Leibler}
\newacronym{VIN}{vin}{value iteration network}
\newacronym{GRU}{gru}{gated recurrent unit}
\newacronym{LSTM}{lstm}{long short term memory}
\newacronym{NN}{nn}{neural network}
\newacronym{SSM}{ssm}{state space model}
\newacronym{ADR-A2C}{adr-a2c}{action-specific deep recurrent A$2$C network}
\newacronym[firstplural=Bayes-adaptive partically observable decision processes]{BA-POMDP}{ba-pomdp}{Bayes-adaptive partically observable decision process}
\newacronym{ESS}{ess}{effective sample size}
\newacronym{BPTT}{bptt}{backpropagation-throught-time}

\newcommand{\given}{\lvert}

\DeclareMathOperator{\E}{\mathbb{E}}
\DeclareMathOperator{\ELBO}{\acrshort{ELBO}}
\DeclareMathOperator{\DVRL}{\acrshort{DVRL}}

\usepackage{paralist}

%% file: include/colors.tex
\usepackage{xcolor}
\colorlet{dark-blue}{blue!50!black}
\colorlet{dark-cyan}{cyan!75!black}
\colorlet{dark-purple}{purple!50!black}
\colorlet{dark-red}{red!75!black}
\colorlet{dark-green}{green!75!black}
\colorlet{dark-orange}{orange!50!black}
\colorlet{dark-gray}{black!75}
\colorlet{light-gray}{black!30}
\colorlet{hidden}{light-gray}
\colorlet{todo}{red!85!black}
\colorlet{todoref}{purple!70!black}
\definecolor{nice-red}{HTML}{E41A1C}
\definecolor{nice-orange}{HTML}{FF7F00}
\definecolor{nice-yellow}{HTML}{FFC020}
\definecolor{nice-green}{HTML}{4DAF4A}
\definecolor{nice-blue}{HTML}{377EB8}
\definecolor{nice-purple}{HTML}{984EA3}

%% file: include/todo.tex
\input{include/colors}

%% file: core/1_intro.tex
% !tex root=../main.tex

\section{Introduction}

Most deep \gls{RL} methods assume that the state of
the environment is fully observable at every time step. However,
this assumption often does not hold in reality, as occlusions and noisy sensors may limit the agent's perceptual abilities.
Such problems can be formalised as \glspl{POMDP} \citep{astrom1965optimal, kaelbling1998planning}.
Because we usually do not have access to the true generative model of our environment, there is a need for reinforcement learning methods that can
tackle \glspl{POMDP} when only a stream of observations is given, without any prior knowledge of the latent state space or the transition and observation functions.

\Glspl{POMDP} are notoriously hard to solve: since the current observation does in general not 
carry all relevant information for choosing an action,
information must be aggregated over time and in general the entire history must be taken into account.

% NOTE: I couldn't use the \gls abbreviation things because this would be the first time we use the abbreviation and the captions become way too long.
\begin{figure}[t]
	\centering
	\begin{subfigure}{\columnwidth}
		\centering
		\includegraphics[width=0.9\columnwidth]{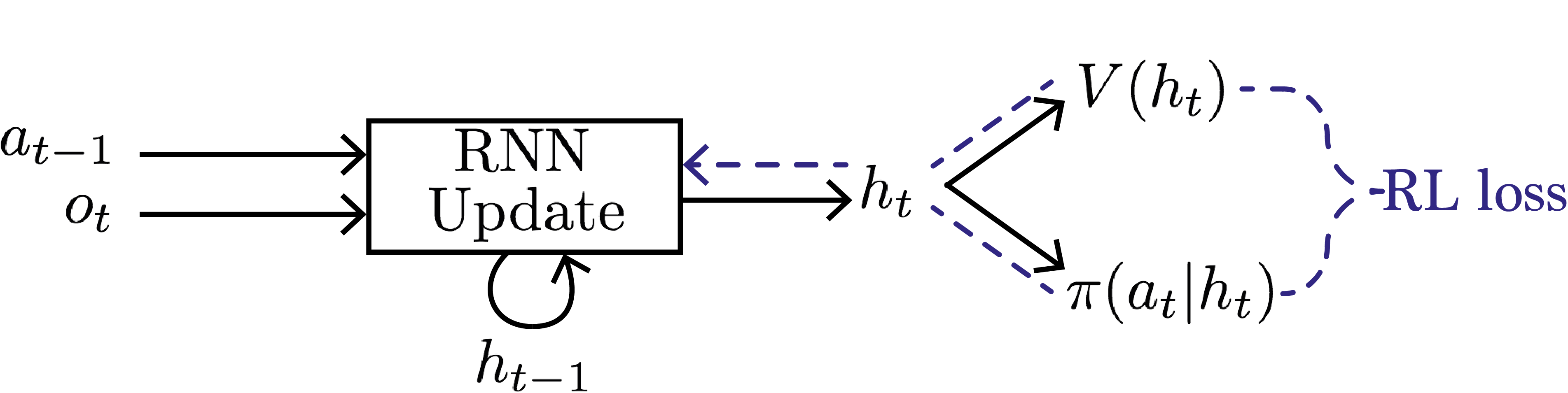}
		\caption{\textbf{RNN-based approach.} The RNN acts as an encoder for the action-observation history, on which actor and critic are conditioned. The networks are updated end-to-end with an RL loss.} \vspace{0.25cm}
		\label{fig:highlevel_rnn}
	\end{subfigure}
	\begin{subfigure}{\columnwidth}
		\centering
		\includegraphics[width=0.9\columnwidth]{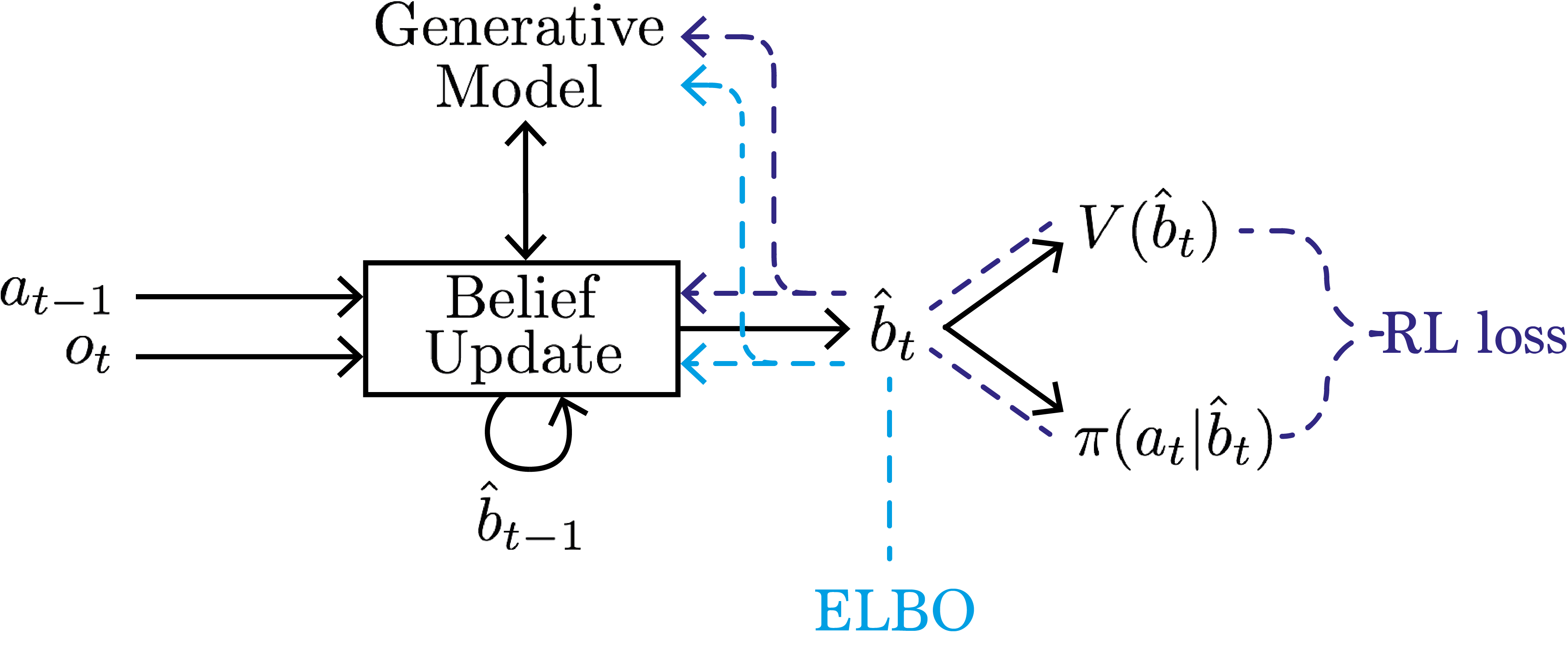}
		\caption{\textbf{DVRL.} The agent learns a generative model which is used
      to update a belief distribution. Actor and critic now condition on the
      belief. The generative model is learned to optimise both the ELBO \emph{and} the RL loss.}
		\label{fig:highlevel_dvrl}
	\end{subfigure}
	\caption{Comparison of \acrshort{RNN} and \acrshort{DVRL} encoders.}
 \label{fig:overview_methods}
\end{figure}

This history can be encoded either by remembering
features of the past \citep{mccallum1993overcoming} or by performing inference to determine the distribution
over possible latent states \citep{kaelbling1998planning}.
However, the computation of this \emph{belief
  state} requires knowledge of the model.

Most previous work in deep learning relies on training a
\gls{RNN} to summarize the past. Examples are the \gls{DRQN}
\citep{hausknecht2015deep} and the \gls{ADRQN} \citep{zhu2017improving}.
Because these approaches are completely model-free, they place a heavy
burden on the \gls{RNN}.
Since performing inference implicitly requires a known or learned model, they
are likely to summarise the history either by only remembering features of the
past or by computing simple heuristics instead of actual belief states.
This is often suboptimal in complex tasks. Generalisation is also often
easier over beliefs than over trajectories since distinct histories can lead to
similar or identical beliefs. 

The premise of this work is that deep policy learning for \glspl{POMDP} can be
improved by taking less of a black box approach than \gls{DRQN} and \gls{ADRQN}.
While we do not want to assume prior knowledge of the transition and observation
functions or the latent state representation, we want to allow the agent to
learn models of them and infer the belief state using this learned model.

To this end, we propose \gls{DVRL}, which implements this approach by providing a helpful inductive bias to the agent. 
In particular, we develop an algorithm that can learn an internal generative
model and use it to perform approximate inference to update the belief state.
Crucially, the generative model is not only learned based on an \gls{ELBO}
objective, but also by how well it enables maximisation of the expected return.
This ensures that, unlike in an unsupervised application of \glspl{VAE}, the
latent state representation and the inference performed on it are suitable for
the ultimate control task. 
Specifically, we develop an approximation to the \gls{ELBO} based on
\gls{AESMC} \citep{le2018autoencoding}, allowing joint optimisation with the
$n$-step policy gradient update.
Uncertainty in the belief state is captured by a particle ensemble.
A high-level overview of our approach in comparison to previous \gls{RNN}-based methods is shown in Figure \ref{fig:overview_methods}.

We evaluate our approach on Mountain Hike and several \emph{flickering} Atari games.
On Mountain Hike, a low dimensional, continuous environment, we can show that \gls{DVRL} is better than an
\gls{RNN} based approach at inferring the required information from past
observations for optimal action selection in a simple setting.
Our results on flickering Atari show that this advantage extends to complex environments with
high dimensional observation spaces.
Here, partial observability is introduced by
\begin{inparaenum}[(1)]
  \item using only a single frame as input at each time step and 
  \item returning a blank screen instead of the true frame with probability 0.5. 
\end{inparaenum}

%% file: core/3_background.tex
% !tex root=../main.tex
\section{Background}

In this section, we formalise \glspl{POMDP} and provide background on
recent advances in \glspl{VAE} that we use. 
Lastly, we describe the policy gradient loss based on
$n$-step learning and \gls{A2C}.

\subsection{Partially Observable Markov Decision
  Processes} \label{sec:background-pomdp}

A partially observable Markov decision process (\gls{POMDP}) is a tuple
$(\mathcal{S}, \mathcal{A}, \mathcal{O}, F, U, R, b_0)$, where $\mathcal{S}$ is
the state space, $\mathcal{A}$ the action space, and $\mathcal{O}$ the observation space. We
denote as $s_t\in\mathcal{S}$ the latent state at
time $t$, and the distribution over initial states $s_0$ as $b_0$, the initial
belief state. When an action
$a_{t}\in\mathcal{A}$ is executed, the state changes according to the
transition distribution, $s_{t+1}\sim F(s_{t+1}|s_t,a_{t})$. Subsequently, the
agent receives a noisy or partially occluded observation $o_{t+1}\in\mathcal{O}$
according to the distribution $o_{t+1}\sim
U(o_{t+1}|s_{t+1},a_{t})$, and a reward $r_{t+1}\in\mathbb{R}$ according to the distribution $r_{t+1}\sim R(r_{t+1}|s_{t+1},a_{t})$.

An agent acts according to its policy $\pi(a_t|o_{\le t}, a_{<t})$ which returns
the probability of taking action $a_t$ at time $t$, and where $o_{\le t}=(o_1,
\dotsc, o_t)$ and $a_{< t}=(a_0, \dotsc, a_{t-1})$ are the observation and
action histories, respectively. The agent's goal is to learn a policy $\pi$ that
maximises the expected future return
\begin{equation}
  \label{eq:J}
  J = \mathbb{E}_{p(\tau)}\left[\sum_{t=1}^T \gamma^{t-1} r_t \right],
\end{equation}
over trajectories $\tau=(s_0,a_0, \dots, a_{T-1},s_T)$ induced by its
policy\footnote{The trajectory length $T$ is stochastic and depends
  on the time at which the agent-environment interaction ends.}, where $0\le\gamma<1$ is
the discount factor.
We follow the convention of setting $a_0$ to \emph{no-op}~\citep{zhu2017improving}.
% We set $a_0$ to a \emph{no-op} as is typical.

In general, a \gls{POMDP} agent must condition its actions on the entire
history $(o_{\le t}, a_{< t})\in\mathcal{H}_t$. The exponential growth in $t$ of
$\mathcal{H}_t$ can be addressed, e.g., with suffix trees
\citep{mccallum1996reinforcement,shani2005model,bellemare2014skip,bellemare2015count,messias2017dynamic}.
However, those approaches suffer from large memory requirements and are
only suitable for small discrete observation spaces. 

Alternatively, it is possible to infer the filtering
distribution $p(s_t|o_{\le t}, a_{< t})=:b_t$, called the \emph{belief state}.
This is a sufficient statistic of the history that can be used as input to an optimal
policy $\pi^\star(a_t|b_t)$. 
The belief space does not grow exponentially, but the belief update step
requires knowledge of the model: 
\begin{equation} \label{eq:bayes-update} 
  \begin{split}
    b_{t+1} & = \frac{\int b_tU(o_{t+1}|s_{t+1},a_{t})F(s_{t+1}|s_{t},a_{t}) \mathrm{d} s_{t}}{\int
    \int b_tU(o_{t+1}|s_{t+1},a_{t})F(s_{t+1}|s_{t},a_{t}) \,\mathrm ds_t
    \,\mathrm ds_{t + 1}}. 
  \end{split}
\end{equation}

\subsection{Variational Autoencoder} \label{sec:background-vae}

We define a family of priors $p_\theta(s)$ over some latent state $s$ and decoders
$p_\theta(o | s)$ over observations $o$, both parameterised by $\theta$.
A \acrlong{VAE} (\gls{VAE}) learns $\theta$ by
maximising the sum of log marginal likelihood terms $\sum_{n = 1}^N \log
p_\theta(o^{(n)})$ for a dataset $(o^{(n)})_{n = 1}^N$ where $p_\theta(o) = \int
p_\theta(o \given s)p_\theta(s) \,\mathrm ds$~\cite{rezende2014stochastic, kingma2014auto}) .
Since evaluating the log marginal likelihood is
intractable, the \gls{VAE} instead maximises a sum of \glspl{ELBO} where each individual
\gls{ELBO} term is a lower bound on the log marginal likelihood,
\begin{align}
  \ELBO(\theta, \phi, o)
  = \E_{q_\phi(s \given o)}\left[ \log \frac{p_\theta(o|s) p_\theta(s)}{q_\phi(s \given o)} \right], \label{eq:elbo-vae} % \\
\end{align}
for a family of encoders $q_\phi(s \given o)$ parameterised by $\phi$. This
objective also forces $q_\phi(s \given o)$ to approximate the posterior
$p_\theta(s \given o)$ under the learned model. Gradients of (\ref{eq:elbo-vae})
are estimated by Monte Carlo sampling with the reparameterisation trick
\citep{kingma2014auto,rezende2014stochastic}. 

\subsection{VAE for Time Series}
\label{sec:background-aesmc}

For sequential data, we assume that a series of latent states $s_{\le
  T}$  gives rise to a series of observations $o_{\le T}$. 
We consider a family of
generative models parameterised by $\theta$ that consists of the initial
distribution $p_\theta(s_0)$, transition distribution $p_\theta(s_t \given s_{t
  - 1})$ and observation distribution $p_\theta(o_t \given s_t)$. Given a family
of encoder distributions $q_\phi(s_t \given s_{t -
  1}, o_t)$, we can also estimate the gradient of the \gls{ELBO} term in the
same manner as in \eqref{eq:elbo-vae}, noting that:
\begin{align}
  p_\theta(s_{\le T}, o_{\le T}) &= p_\theta(s_0) \prod_{t = 1}^T p_\theta(s_t \given s_{t - 1}) p_\theta(o_t \given s_t),   \\
  q_\phi(s_{\le T} \given o_{\le T}) &= p_\theta(s_0) \prod_{t = 1}^T q_\phi(s_t \given s_{t - 1}, o_t),
\end{align}
where we slightly
abuse notation for $q_\phi$ by ignoring the fact that we sample from the model
$p_\theta(s_0)$ for $t=0$.
\citet{le2018autoencoding}, \citet{maddison2017filtering} and \citet{naesseth2017variational} introduce a new
\gls{ELBO} objective based on \gls{SMC} \citep{doucet2009tutorial} that allows
faster learning in time series:
\begin{align}
  \ELBO_{\text{SMC}}(\theta, \phi, o_{\leq T}) = \E\left[\sum_{t = 1}^T\log \left(  \frac{1}{K} \sum_{k = 1}^K w_t^k \right)\right], \label{eq:aesmc-loss}
\end{align}
where $K$ is the number of particles and $w_t^k$ is the weight of particle $k$ at time $t$.
Each particle is a tuple containing a weight $w_t^k$ and a value $s_t^k$ which
is obtained as follows.
Let $s_0^k$ be samples from $p_\theta(s_0)$ for $k = 1, \dotsc, K$.
For $t = 1, \dotsc, T$, the weights $w_t^k$ are obtained by resampling the
particle set $(s_{t - 1}^k)_{k = 1}^K$ proportionally to the previous weights and computing 
\begin{equation}
\label{eq:aesmc-weights}
w_t^k = \frac{p_\theta(s_t^k \given s_{t - 1}^{u_{t - 1}^k}) p_\theta(o_t \given s_t^k)} {q_\phi(s_t^k \given s_{t - 1}^{u_{t - 1}^k}, o_t)},
\end{equation}
where $s_t^k$ corresponds to a value sampled from $q_\phi(\cdot \given s_{t - 1}^{u_{t - 1}^k}, o_t)$ and $s_{t - 1}^{u_{t - 1}^k}$ corresponds to the resampled particle with the ancestor index $u_0^k = k$ and $u_{t - 1}^k \sim \mathrm{Discrete}((w_{t - 1}^k / \sum_{j = 1}^K w_{t - 1}^j)_{k = 1}^K)$ for $t = 2, \dotsc, T$.

\subsection{A2C} \label{sec:background-a2c} 
One way to learn the parameters $\rho$ of an agent's policy $\pi_\rho(a_t|s_t)$ is to use
$n$-step learning with
\gls{A2C}~\citep{dhariwal2017openaibaselines,wu2017scalable}, the synchronous
simplification of \gls{A3C}~\citep{mnih2016asynchronous}. An actor-critic approach
can cope with continuous actions and avoids the need to draw
state-action sequences from a replay buffer. 
The method proposed in this paper is however equally applicable to other deep \gls{RL} algorithms.

For $n$-step learning, starting at time $t$, the current policy performs
$n_s$ consecutive steps in $n_e$ parallel environments. 
The gradient update is based on this mini-batch of size $n_e\times n_s$.
The target for the value-function $V_\eta(s_{t+i}), i = 0, \dotsc,
n_{s}-1$, parameterised by $\eta$, is the appropriately discounted sum of on-policy rewards up until time
$t+n_s$ and the off-policy bootstrapped value $V_\eta^-(s_{t+n_s})$. The minus
sign denotes that no gradients are propagated through this value.
Defining the advantage function as
\begin{align}
\begin{split}
    A_\eta^{t,i}(s_{t+i},a_{t+i}) := & \left(\sum_{j = 0}^{n_s - i - 1} \gamma^{j} r_{t + i + j } \right) \\
    & +  \gamma^{n_s - i} V_\eta^-(s_{t + n_s}) - V_\eta(s_{t+i}) \ ,
\end{split}
\end{align}
the \acrshort{A2C} loss for the policy parameters $\rho$ at time $t$ is
\begin{align}
\begin{split}
  \mathcal L_t^A&(\rho) = -\frac{1}{n_en_s}\\
  & \sum_{\text{envs}}^{n_e} \sum_{i = 0}^{n_s - 1} \log \pi_{\rho}(a_{t+i} | s_{t+i}) A_{\eta}^{t,i,-}(s_{t+i},a_{t+i}).
    \label{eq:LA}
\end{split}
\end{align}
and the value function loss to learn $\eta$ can be written as
\begin{align}
  \mathcal L_t^V(\eta) = \frac{1}{n_en_s}\sum_{\text{envs}}^{n_e} \sum_{i = 0}^{n_s - 1} A^{t,i}_\eta(s_{t+i},a_{t+i})^2 .
    \label{eq:LV}
\end{align}

Lastly, an entropy loss is added to encourage exploration:
\begin{align}
  \mathcal L_t^H(\rho) = -\frac{1}{n_en_s}\sum_{\text{envs}}^{n_e} \sum_{i = 0}^{n_s - 1} H(\pi_\rho(\cdot | s_{t+i})),
    \label{eq:LH}
\end{align}
where $H(\cdot)$ is the entropy of a distribution.

%% file: core/4_methods.tex
% !tex root=../main.tex

\section{Deep Variational Reinforcement Learning}
\label{sec:Method}

Fundamentally, there are two approaches to aggregating the history in the presence of partial observability: remembering features of the past or maintaining beliefs.

In most previous work, including \gls{ADRQN} \citep{zhu2017improving},
the current history $(a_{\le t}, o_{<t})$ is encoded by an \gls{RNN}, which
leads to the recurrent update equation for the latent state $h_{t}$:  
\begin{equation}
  \label{eq:rnn-update}
  h_{t} = \mathrm{RNNUpdate}_\phi(h_{t-1}, a_{t-1}, o_{t}) \ .
\end{equation}
Since this approach is model-free, it is unlikely to approximate belief update
steps, instead relying on memory or simple heuristics.

Inspired by the premise that a good way to
solve many \glspl{POMDP} involves \begin{inparaenum}[(1)]
  \item estimating the transition
  and observation model of the environment, 
  \item performing inference under this
  model, and 
  \item choosing an action based on the inferred belief state,
\end{inparaenum}
we propose \glsreset{DVRL}\gls{DVRL}.
It extends the \gls{RNN}-based approach to explicitly support belief inference.
Training everything end-to-end shapes the learned model to be useful for
the \acrshort{RL} task at hand, and not only for predicting observations.

We first explain our baseline architecture and training
method in Section \ref{sec:baseline}. For a fair comparison, we modify the original
architecture of \citet{zhu2017improving} in several ways. We find that our new
baseline outperforms their reported results in the majority of cases. 

In Sections \ref{sec:extending} and \ref{sec:recurrent-update}, we explain
our new latent belief state $\hat{b}_t$ and the recurrent update
function
\begin{equation}
\hat{b}_{t}=\mathrm{BeliefUpdate}_{\theta,\phi}(\hat{b}_{t-1}, a_{t-1}, o_{t})
\end{equation}
which replaces \eqref{eq:rnn-update}.
Lastly, in Section \ref{sec:loss}, we describe our modified loss function, which allows 
learning the model jointly with the policy.

\subsection{Improving the Baseline}
\label{sec:baseline}

While previous work often used $Q$-learning to train the policy
\citep{hausknecht2015deep,zhu2017improving,foerster2016learning,narasimhan2015language},
we use $n$-step A2C.
This avoids drawing entire trajectories from a replay buffer and allows
continuous actions.

Furthermore, since A2C interleaves unrolled trajectories and performs a parameter
update only every $n_s$ steps, it makes it feasible to maintain an
approximately correct latent state. A small bias is introduced by not
recomputing the latent state after each gradient update step.

We also modify the implementation of \gls{BPTT} for $n$-step A2C in the case of policies with latent
states. Instead of backpropagating gradients only through the computation graph
of the current update involving $n_s$ steps, we set the size of the computation
graph independently to involve $n_g$ steps. This leads to an average \gls{BPTT}-length of
$(n_g -1)/2$.\footnote{This is implemented in
  PyTorch using the \texttt{retain\_graph=True} flag in the \texttt{backward()} function.} 
This decouples the bias-variance tradeoff
of choosing $n_s$ from the bias-runtime tradeoff of choosing $n_g$.
Our experiments show that choosing $n_g>n_s$ greatly improves the agent's performance.

\subsection{Extending the Latent State}
\label{sec:extending}

For \gls{DVRL}, we extend the latent state to be a set of $K$ particles,
capturing the uncertainty in the belief state \citep{thrun2000monte,silver2010monte}.
Each particle consists of the triplet $(h_t^k, z_t^k,
w_t^k)$ \citep{chung2015recurrent}.
The value $h_t^k$ of particle $k$ is the latent state of an \gls{RNN}; $z_t^k$
is an additional stochastic latent state that allows us to learn stochastic transition
models; and $w_t^k$ assigns each particle an importance weight.

Our latent state $\hat{b}_t$ is thus an approximation of the belief state
in our learned model
\begin{multline}
    p_\theta(h_{\le T}, z_{\le T}, o_{\le T} \given a_{< T}) = p_\theta(h_0)\prod_{t=1}^T p_\theta(z_t|h_{t-1},a_{t-1}) \\
     p_\theta(o_t|h_{t-1},z_t, a_{t-1}) \delta_{\psi^{\text{RNN}}_\theta(h_{t-1}, z_t, a_{t-1}, o_t)}(h_t),
\end{multline}
with stochastic transition model $p_\theta(z_t|h_{t-1},a_{t-1})$, decoder $p_\theta(o_t|h_{t-1},z_t, a_{t-1})$,
and deterministic transition function $h_t = \psi^{\text{RNN}}_\theta(h_{t-1},
z_t, a_{t-1}, o_t)$ which is denoted using the delta-distribution $\delta$ and
for which we use an \gls{RNN}.
The model is trained to jointly optimise the \gls{ELBO} and the expected return. 

\subsection{Recurrent Latent State Update}
\label{sec:recurrent-update}

\begin{figure*}
	\includegraphics[width=\textwidth]{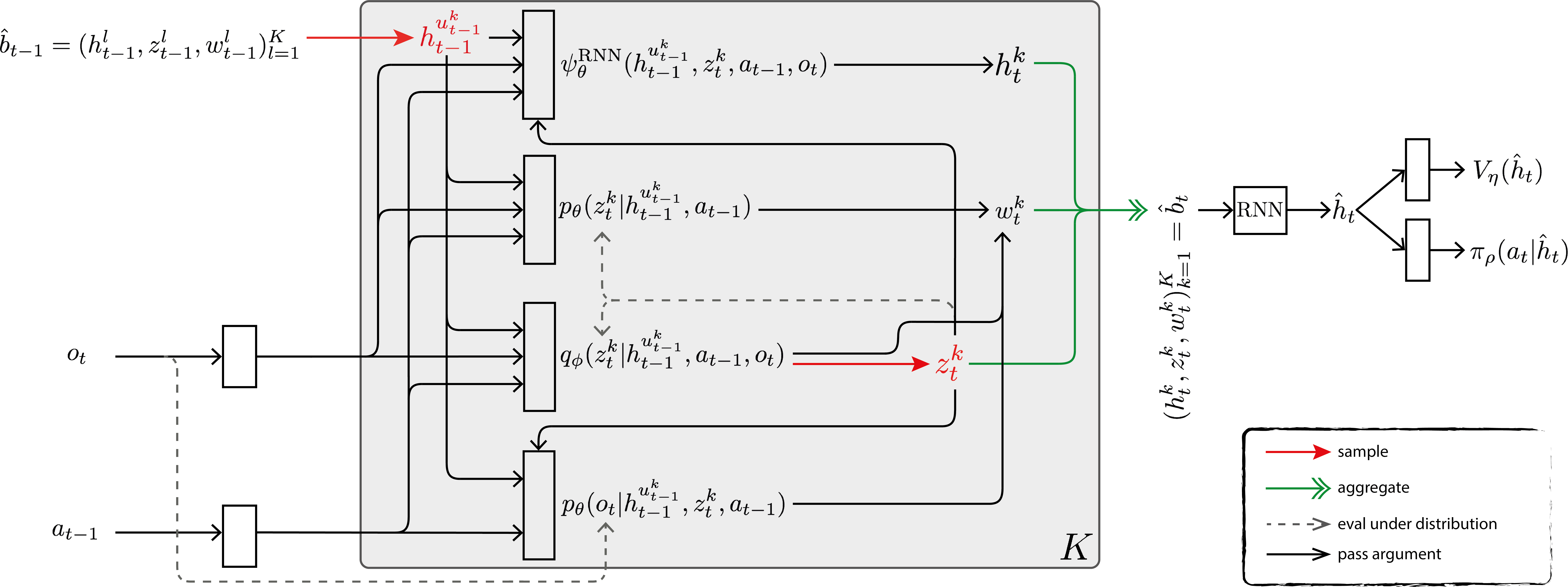}
	\caption{Overview of \gls{DVRL}. We do the following $K$ times to compute our new
    belief $\hat{b}_t$:
    Sample an ancestor index $u_{t-1}^k$ based on the previous weights $w_{t-1}^{1:K}$ (Eq.
    \ref{eq:dvrl-u}).
    Pick the ancestor particle value $h^{u_{t-1}^k}_{t-1}$ and use it to
    sample a new stochastic latent state $z_t^k$ from the encoder
    $q_\phi$ (Eq. \ref{eq:dvrl-z}). 
    Compute $h_t^k$ (Eq. \ref{eq:dvrl-h}) and $w_t^k$ (Eq. \ref{eq:dvrl-w}).
    Aggregate all $K$ values into the new belief $\hat{b}_t$ and summarise them into
    a vector representation $\hat{h}_t$ using a second \gls{RNN}.
    Actor and critic can now condition on $\hat{h}_t$ and $\hat{b}_t$ is used as
    input for the next timestep.
    Red arrows denote random sampling, green arrows the aggregation of $K$
    values. Black solid arrows denote the passing of a value as argument to a function and black dashed ones
    the evaluation of a value under a distribution. Boxes indicate neural
    networks.
    Distributions are normal or Bernoulli distributions whose parameters are outputs of the neural network.} 
  \label{fig:dvp}
\end{figure*}

To update the latent state, we proceed as follows:
\begin{align}
u_{t-1}^k & \sim \mathrm{Discrete}\left(\frac{w_{t-1}^k}{\sum_{j=1}^K w_{t-1}^j}\right) \label{eq:dvrl-u}\\
z_{t}^k & \sim q_\phi(z_{t}^k|h_{t-1}^{u_{t-1}^k}, a_{t-1}, o_{t}) \label{eq:dvrl-z}\\
h_{t}^k & = \psi_\theta^{\text{RNN}}(h_{t-1}^{u_{t-1}^k}, z_{t}^k, a_{t-1}, o_{t}) \label{eq:dvrl-h}\\
  w_{t}^k & = \frac{
	p_\theta(z_{t}^k \given h_{t-1}^{u_{t-1}^k}, a_{t-1}) 
	p_\theta(o_{t} \given h_{t-1}^{u_{t-1}^k}, z_{t}^k, a_{t-1})}
	{q_\phi(z_{t}^k|h_{t-1}^{u_{t-1}^k}, a_{t-1}, o_{t})} \label{eq:dvrl-w} \ .
\end{align}
First, we resample particles based on their weight by drawing ancestor
indices $u_{t-1}^k$. This improves model
learning \citep{le2018autoencoding,maddison2017filtering} and allows us to
train the model jointly with the $n$-step loss (see
Section \ref{sec:loss}). 

For $k=1\dots K$, new values for $z_{t}^k$ are sampled from the encoder 
$q_\phi(z_{t}^k|h_{t-1}^{u_{t-1}^k}, a_{t-1}, o_{t})$ which conditions on the resampled ancestor values
$h_{t-1}^{u_{t-1}^k}$ as well as the last actions $a_{t-1}$ and current observation $o_t$. 
Latent variables $z_t$ are sampled using the reparameterisation trick.
The values $z_{t}^k$, together with $h_{t-1}^{u_{t-1}^k}, a_{t-1}$ and $o_t$, are then passed to the transition function
$\psi^{\text{RNN}}_\theta$ to compute $h_{t}^k$.

The weights $w_{t}^k$ measure how likely each new latent state value
$(z_{t}^k, h_{t}^k)$ is under the model and how well it explains the current observation.

To condition the policy on the belief $\hat{b}_{t}=(z_{t}^k, h_{t}^k,w_{t}^k)_{k=1}^K$, we need to encode the set
of particles into a vector representation $\hat{h}_t$.
We use a second \gls{RNN} that sequentially takes in
each tuple $(z_{t}^k, h_{t}^k,w_{t}^k)$ and its last latent state is $\hat{h}_t$.

Additional encoders are used for $a_t$, $o_t$ and $z_t$; see Appendix
\ref{sec:ap:experiments} for details. 
Figure \ref{fig:dvp} summarises the entire update step.

\subsection{Loss Function}
\label{sec:loss}

To encourage learning a model, we include the term
\begin{align}
  \mathcal L_t^{\ELBO}(\theta, \phi) = -\frac{1}{n_en_s}\sum_{\text{envs}}^{n_e} \sum_{i = 0}^{n_s - 1} \log\left(\frac{1}{K} \sum_{k = 1}^K w_{t + i}^k \right)  \label{eq:elbo-loss}
\end{align}
in each gradient update every $n_s$ steps. This leads to the overall loss:
\begin{align}
  \mathcal L_t^{\DVRL}(\rho, \eta, \theta, \phi) = \mathcal L_t^A(\rho, \theta, \phi) + \lambda^H \mathcal L_t^H(\rho, \theta, \phi) + \nonumber\\ \lambda^V \mathcal L_t^V(\eta, \theta, \phi) + \lambda^{E} \mathcal L_t^{\ELBO}(\theta, \phi) \label{eq:dvp-loss} \ .
\end{align}
Compared to \eqref{eq:LA}, \eqref{eq:LV} and \eqref{eq:LH}, the losses now also
depend on the encoder parameters $\phi$ and, for \gls{DVRL}, model parameters
$\theta$, since the policy and value function now condition on the
latent states instead of $s_t$. 
By introducing the $n$-step approximation $\mathcal{L}_t^{\text{ELBO}}$, we can
learn $\theta$ and $\phi$ to jointly optimise the \gls{ELBO} and the \gls{RL} loss
$\mathcal{L}^A_t+\lambda^H\mathcal{L}^H_t+\lambda^V\mathcal{L}^V_t$.

If we assume that observations and actions are drawn from the stationary state distribution
induced by the policy $\pi_\rho$, then $\mathcal{L}^{\text{ELBO}}_t$ is a
stochastic approximation to the action-conditioned \gls{ELBO}:
\begin{multline}
  \frac{1}{T}\mathbb{E}_{p(\tau)} \ELBO_\text{SMC}(o_{\le T}|a_{< T}) =  \\
  \frac{1}{T}\mathbb{E}_{p(\tau)} \mathbb{E}\left[\left. \sum_{t = 1}^{T} \log\left(\frac{1}{K} \sum_{k =
          1}^K w_{t}^k \right)\right| a_{\le T}\right] \label{eq:total-elbo} \ ,
\end{multline}
which is a conditional extension of (\ref{eq:aesmc-loss}) similar to the extension of \glspl{VAE}
by \citet{sohn2015learning}.  
The expectation over $p(\tau)$ is approximated by sampling trajectories and
the sum $\sum_{t=1}^T$ over the entire trajectory is approximated by a sum $\sum_{i=t}^{t+n_s-1}$ over only a part of it.

The importance of the resampling step (\ref{eq:dvrl-u}) in allowing this approximation becomes clear
if we compare (\ref{eq:total-elbo}) with the \gls{ELBO} for the \gls{IWAE} that
does not include resampling \citep{doucet2009tutorial,burda2016importance}:
\begin{equation}
\label{eq:iwae}
  \ELBO_{\text{IWAE}}(o_{\le T} | a_{<T}) = \E\left[\left. \log \left( \frac{1}{K} \sum_{k = 1}^K \prod_{t = 1}^T w_t^k \right)\right| a_{\le T}\right] \ .
\end{equation}
Because this loss is not additive over time, we cannot approximate it with shorter
parts of the trajectory.

%% file: core/2_related-work.tex
% !tex root=../main.tex

\section{Related Work}

Most existing \gls{POMDP} literature focusses on \emph{planning}
algorithms, where the transition and observation functions, as well as a representation of the latent
state space, are known \citep{barto1995learning,
  mcallester1999approximate, pineau2003point,
  ross2008online, oliehoek2008exploiting, roijers2015point}.
In most realistic domains however, these are not known a priori.

There are several approaches that utilise \glspl{RNN} in \glspl{POMDP}
\citep{bakker2002reinforcement, wierstra2007solving,zhang2015policy,heess2015memory},
including multi-agent settings \citep{foerster2016learning}, learning text-based
fantasy games \citep{narasimhan2015language} or, most recently, applied to
Atari \citep{hausknecht2015deep,zhu2017improving}.
As discussed in Section \ref{sec:Method}, our algorithm extends those approaches
by enabling the policy to explicitly reason about a model and the belief state.

Another more specialised approach called QMDP-Net \citep{karkus2017qmdp} learns a \gls{VIN} \citep{tamar2016value} end-to-end and uses it as a transition model for planning. However, a \gls{VIN} makes strong assumptions about the transition function and in QMDP-Net the belief update must be performed analytically. 

The idea to learn a particle filter based policy that is trained using policy
gradients was previously proposed by \citet{coquelin2009particle}. However, they
assume a known model and rely on finite differences for gradient estimation. 

Instead of optimising an \gls{ELBO} to learn a maximum-likelihood approximation for the latent representation and corresponding transition and
observation model, previous work also tried to learn those dynamics using
spectral methods \citep{azizzadenesheli2016reinforcement}, a Bayesian
approach \citep{ross2011bayesian,katt2017learning}, or nonparametrically \citep{doshi2015bayesian}.
However, these approaches do not scale to large or continuous state and observation spaces.
For continuous states, actions, and observations with Gaussian noise, a Gaussian process model can be learned \citep{deisenroth2012solving}.
An alternative to learning an (approximate) transition and observation model is
to learn a model over trajectories \citep{willems1995context}. However, this is
again only possible for small, discrete observation spaces.

Due to the complexity of the learning in \glspl{POMDP}, previous work
already found benefits to using auxiliary losses. Unlike the losses proposed by \citet{lample2017playing}, we do not require additional information from the environment.
The \emph{UNREAL} agent \citep{jaderberg2016reinforcement} is, similarly to our work, motivated by the idea to improve the latent representation by utilising all the information already obtained from the environment.
While their work focuses on finding unsupervised auxiliary losses that provide good training signals, our goal is to use the auxiliary loss to better align the network computations with the task at hand by incorporating prior knowledge as an inductive bias.

There is some evidence from recent experiments on the dopamine system in mice
\citep{babayan2018belief} showing that their response to
ambiguous information is consistent with a theory operating on belief states.

%% file: core/5_experiments.tex
% !tex root=../main.tex

\section{Experiments}
\label{sec:experiments}

We evaluate \gls{DVRL} on Mountain Hike and on flickering Atari.  
We show that \gls{DVRL} deals better with noisy or partially
occluded observations and that this scales to high dimensional and continuous
observation spaces like images and complex tasks. We also perform a series of ablation studies,
showing the importance of using many particles, including the \gls{ELBO}
training objective in the loss function, and jointly optimising the \gls{ELBO}
and RL losses.

More details about the environments and model architectures can be found in
Appendix \ref{sec:ap:experiments} together with additional results and visualisations.
All plots and reported results are smoothed over time and
parallel executed environments. We average over five random
seeds, with shaded areas indicating the standard deviation.

\subsection{Mountain Hike}
\begin{figure}[!htb]
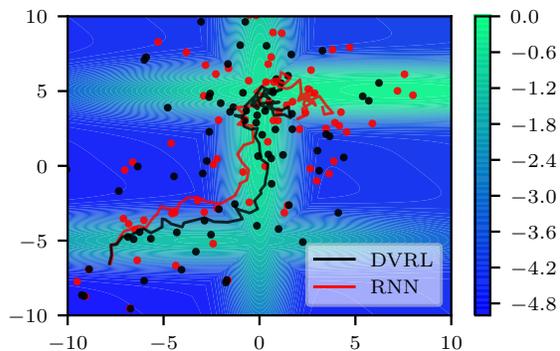

  \centering
  \includegraphics[width=0.95\linewidth]{{{images/Experiments/ForRudolf}}}
  \caption{Mountain Hike is a continuous control task with observation noise $\sigma_{o}=3$.
    Background colour indicates rewards. Red line: trajectory for \gls{RNN} based encoder.
    Black line: trajectory for \gls{DVRL} encoder. Dots:
    received observations. Both runs share the same noise values $\epsilon_{i,t}$.
   \gls{DVRL} achieves higher returns (see Fig.\
   \ref{fig:hike-plots}) by better estimating its current
   location and remaining on the high reward mountain ridge.}
  \label{fig:hike-env}
\end{figure}

In this task, the agent has to navigate along a mountain ridge, but only receives
noisy measurements of its current location. Specifically, we have
$\mathcal{S} = \mathcal{O} = \mathcal{A} = \mathbb{R}^2$ where $s_t=[x_t,y_t]^T\in\mathcal{S}$
and $o_t=[\hat{x}_t,\hat{y}_t]^T\in\mathcal{O}$ are true and observed
coordinates respectively and $a_t = [\Delta x_t, \Delta y_t]^T\in\mathcal{A}$
is the desired step. 
Transitions are given by $s_{t+1}= s_{t} + \tilde{a}_t + \epsilon_{s,t}$ with
$\epsilon_{s,t}\sim\mathcal{N}(\cdot|0, 0.25 \cdot I)$ and $\tilde{a}_t$ is the
vector $a_t$ with length capped to $\|\tilde{a}_t\|\le 0.5$.
Observations are noisy with $o_t=s_t+\epsilon_{o,t}$ with
$\epsilon_{o,t}\sim\mathcal{N}(\cdot|0, \sigma_{o}\cdot I)$ and $\sigma_o\in
\{0, 1.5, 3\}$.
The reward at each timestep is $R_t = r(x_t, y_t) - 0.01 \|a_t\|$ where $r(x_t,y_t)$ is shown
in Figure \ref{fig:hike-env}. The starting position is sampled from
$s_0\sim\mathcal{N}(\cdot|[-8.5, -8.5]^T, I)$ and each episode ends
after 75 steps.

\gls{DVRL} used 30 particles and we set
$n_g=25$ for both \gls{RNN} and \gls{DVRL}. The latent state $h$ for the
\gls{RNN}-encoder architecture was of dimension 
256 and 128 for both $z$ and $h$ for \gls{DVRL}.
Lastly, $\lambda^E=1$ and $n_s=5$ were used, together with RMSProp with a learning rate of $10^{-4}$
for both approaches.

\begin{figure}[!htb]
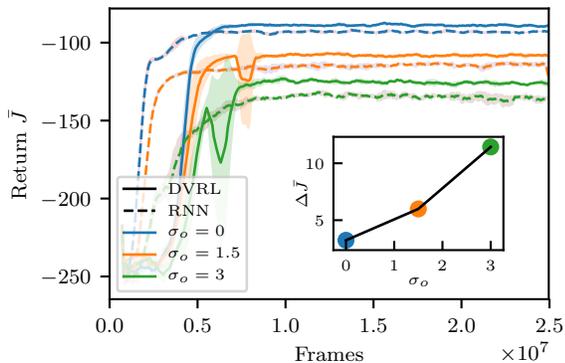

  \centering
  \includegraphics[width=0.95\linewidth]{{{images/Experiments/DeathValley-v0-result.true-}}}
  \caption{Returns achieved in Mountain Hike. Solid lines: \gls{DVRL}.
    Dashed lines: \gls{RNN}. Colour: Noise levels.
    \emph{Inset:} Difference in performance between \gls{RNN} and \gls{DVRL} for same
    level of noise: $\Delta \bar{J}(\sigma_{o}) = \bar{J}(\text{DVRL}, \sigma_o) -
    \bar{J}(\text{RNN}, \sigma_o)$. \gls{DVRL} achieves slighly higher returns for the
    fully observable case and, crucially, its performance deteriorates more slowly
    for increasing observation noise, showing the advantage of \gls{DVRL}'s
    inference computations in encoding the history in the presence of
    observation noise.}
  \label{fig:hike-plots}
\end{figure}

The main difficulty in Mountain Hike is to 
correctly estimate the current position. Consequently, the achieved return
reflects the capability of the network to do so. 
\gls{DVRL} outperforms \gls{RNN} based policies, especially for higher levels of
observation noise $\sigma_o$ (Figure \ref{fig:hike-plots}). 
In Figure \ref{fig:hike-env} we compare the different trajectories for \gls{RNN}
and \gls{DVRL} encoders for the same noise, i.e.
$\epsilon_{s,t}^{\text{RNN}}=\epsilon_{s,t}^{\text{DVRL}}$ and 
$\epsilon_{o,t}^{\text{RNN}}=\epsilon_{o,t}^{\text{DVRL}}$ for all $t$ and $\sigma_o=3$.
\gls{DVRL} is better able to follow the mountain
ridge, indicating that its inference based history aggregation is superior to a
largely memory/heuristics based one.

The example in Figure \ref{fig:hike-env} is representative but selected for
clarity: The shown trajectories have $\Delta J(\sigma_o=3) = 20.7$ compared to an average
value of $\Delta \bar{J}(\sigma_o=3)=11.43$ (see Figure \ref{fig:hike-plots}).

\subsection{Atari}
\label{sec:exp:atari}

We chose flickering Atari as evaluation benchmark, since it was previously used
to evaluate the performance of \gls{ADRQN} \citep{zhu2017improving} and
\gls{DRQN} \citep{hausknecht2015deep}. Atari environments
\citep{bellemare2013arcade} provide a wide set of challenging tasks with high
dimensional observation spaces. 
We test our algorithm on the same subset of games on which \gls{DRQN} and
\gls{ADRQN} were evaluated.

% Describe the environments
Partial observability is introduced by \emph{flickering}, i.e., by a probability
of $0.5$ of returning a blank screen instead of the actual observation.
Furthermore, only one frame is used as the observation.
This is in line with previous work \citep{hausknecht2015deep}. 
We use a frameskip of four\footnote{A frameskip of one is used for Asteroids due to
known rendering issues with this environment} and for the stochastic Atari environments there is
a $0.25$ chance of repeating the current action for a second time at each transition.

\gls{DVRL} used 15 particles and we set $n_g=50$ for both agents.
The dimension of $h$ was 256 for both architectures, as was the dimension of $z$.
Larger latent states decreased the performance for the \gls{RNN} encoder.
Lastly, $\lambda^E=0.1$ and $n_s=5$ was used with a learning rate of $10^{-4}$ for
\gls{RNN} and $2\cdot10^{-4}$ for \gls{DVRL}, selected out of a set of 6 different rates
based on the results on ChopperCommand.

Table \ref{table:win-table} shows the results for the more challenging
stochastic, flickering environments. Results for the deterministic environments,
including returns reported for \gls{DRQN} and \gls{ADRQN}, can be found in
Appendix \ref{sec:ap:experiments}.
\gls{DVRL} significantly outperforms the \gls{RNN}-based policy on five out of ten games and narrowly underperforms
significantly on only one.
This shows that \gls{DVRL} is viable for high dimensional observation spaces with 
complex environmental models.

\begin{table}[!htb]
  \caption{Returns on stochastic and flickering Atari environments,
    averaged over 5 random seeds.
    Bold numbers indicate statistical significance at the 5\% level.
    Out of ten games, \gls{DVRL} significantly outperforms the baseline on five games and
  underperforms narrowly on only one game. Comparisons against \gls{DRQN} and
  \gls{ADRQN} on deterministic Atari environments are in Appendix \ref{sec:ap:experiments}.}
  \label{table:win-table}
  \begin{tabular}{lll}
    \toprule
    Env       & \gls{DVRL}$(\pm std)$          & \gls{RNN}$(\pm std)$   \\
    \midrule
    Pong      & $\bm{18.17} (\pm 2.67)$                & $6.33 (\pm3.03)$           \\
    Chopper   & $\bm{6602} (\pm 449)$                  & $5150 (\pm 488)$           \\
    MsPacman  & $2221 (\pm 199)$                       & $2312 (\pm 358)$           \\
    Centipede & $4240 (\pm 116)$                       & $ 4395 (\pm 224)$          \\
    BeamRider & $1663 (\pm 183)$                       & $1801 (\pm 65)$            \\
    Frostbite & $\bm{297} (\pm 7.85)$                  & $254 (\pm 0.45)$           \\
    Bowling   & $29.53 (\pm 0.23)$                     & $\bm{30.04} (\pm 0.18)$    \\
    IceHockey & $\bm{-4.88} (\pm 0.17)$                & $ -7.10 (\pm 0.60)$        \\
    DDunk     & $\bm{-5.95} (\pm 1.25)$                & $ -15.88 (\pm 0.34)$       \\
    Asteroids & $1539 (\pm 73)$                        & $ 1545(\pm 51)$            \\
    \bottomrule
  \end{tabular}
\end{table}

% Describe the results

\subsection{Ablation Studies}
\begin{figure*}
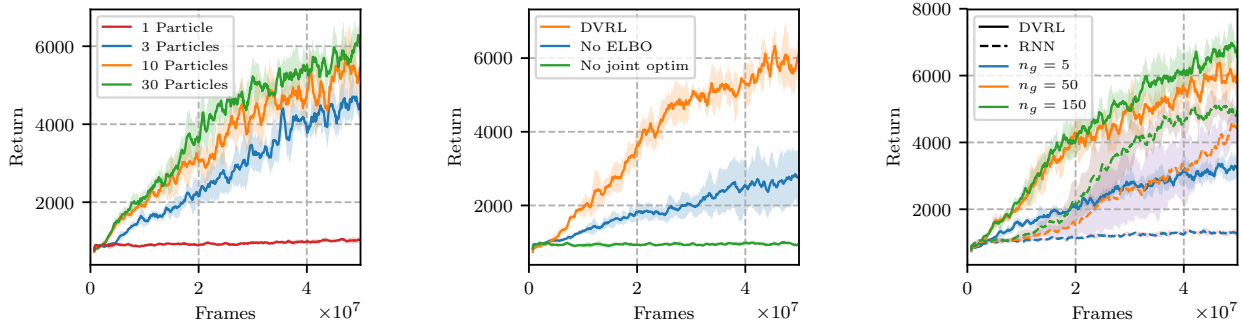

  \begin{subfigure}[t]{0.32\textwidth}
  \includegraphics[width=0.95\textwidth]{{{images/Experiments/ChopperCommandNoFrameskip-v4-result.true-Particle-Numbers}}}
    \caption{Influence of the particle number on performance for \gls{DVRL}.
      Only using one particle is not sufficient to encode enough information in the
      latent state.}
  \end{subfigure}\hfill
  \begin{subfigure}[t]{0.32\textwidth}
  \includegraphics[width=0.95\textwidth]{{{images/Experiments/ChopperCommandNoFrameskip-v0-result.true-Ablation}}}
    \caption{Performance of the full \gls{DVRL} algorithm compared to setting
      $\lambda^E=0$ ("No ELBO") or not backpropagating the policy gradients
      through the encoder ("No joint optim").}
  \end{subfigure}\hfill
  \begin{subfigure}[t]{0.32\textwidth}
  \includegraphics[width=0.95\textwidth]{{{images/Experiments/ChopperCommandNoFrameskip-v4-result.true-Backprop-Length}}}
    \caption{Influence of the maximum backpropagation length $n_g$ on performance. Note
      that \gls{RNN} suffers most from very short lengths. This is consistent
      with our conjecture that \gls{RNN} relies mostly on memory, not inference.}
  \end{subfigure}
  \caption{Ablation studies on flickering ChopperCommand (Atari).}
  \label{fig:ablation}
\end{figure*}

Using more than one particle is important to accurately approximate the belief
distribution over the latent state $(z, h)$.
Consequently, we expect that higher particle numbers provide better information
to the policy, leading to higher returns.
Figure \ref{fig:ablation}a shows that this is indeed the case. 
This is an important result for our architecture, as it also
implies that the resampling step is necessary, as detailed in Section
\ref{sec:loss}. Without resampling, we cannot approximate the \gls{ELBO} on only
$n_s$ observations.

Secondly, Figure \ref{fig:ablation}b shows that the inclusion of
$\mathcal{L}^{\text{ELBO}}$ to encourage model learning is required for good
performance.
Furthermore, not backpropagating the policy gradients through the encoder and only
learning it based on the \gls{ELBO} (``No joint optim") also deteriorates performance.

Lastly, we investigate the influence of the backpropagation length $n_g$ on both
the \gls{RNN} and \gls{DVRL} based policies. While increasing $n_g$ universally
helps, the key insight here is that a short length $n_g=5$ (for an average
\gls{BPTT}-length of 2 timesteps) has a stronger
negative impact on \gls{RNN} than on \gls{DVRL}. This is consistent with our
notion that \gls{RNN} is mainly performing memory based reasoning, for which
longer backpropagation-through-time is required:
The belief update \eqref{eq:bayes-update} in \gls{DVRL}
is a one-step update from $b_t$ to $b_{t+1}$,
without the need to condition on past actions and observations.
The proposal distribution can benefit from extended backpropagation lengths, but
this is not necessary.  
Consequently, this result supports our notion that \gls{DVRL} relies
more on inference computations to update the latent state.

%% file: core/7_conclusion.tex
% !tex root=../main.tex

\section{Conclusion}

In this paper we proposed \gls{DVRL}, a method for solving \glspl{POMDP} given only a
stream of observations, without knowledge of the latent state space or the
transition and observation functions operating in that space. Our method leverages 
a new \gls{ELBO}-based auxiliary loss and incorporates an inductive bias into the
structure of the policy network, taking advantage of our prior knowledge that
an inference step is required for an optimal solution.

We compared \gls{DVRL} to an \gls{RNN}-based architecture and found that we
consistently outperform it on a diverse set of tasks, including a number of Atari
games modified to have partial observability and stochastic transitions.

We also performed several ablation studies showing the necessity of using an
ensemble of particles and of joint optimisation of the \gls{ELBO} and RL objective. 
Furthermore, the results support our claim that the latent state in \gls{DVRL}
approximates a belief distribution in a learned model.

Access to a belief distribution opens up several interesting research
directions.
Investigating the role of better generalisation capabilities and the more powerful
latent state representation on the policy performance of \gls{DVRL} can give
rise to further improvements. 
\gls{DVRL} is also likely to benefit from more powerful model
architectures and a disentangled latent state. 
Furthermore, uncertainty in the belief state and access to a learned model
can be used for curiosity driven exploration in environments with sparse rewards.

%% file: core/8_appendix_experiments.tex
\section{Experiments}
\label{sec:ap:experiments}
\subsection{Implementation Details}

In our implementation, the transition and proposal distributions $p_\theta(z_{t} \given h_{t-1},
a_{t-1})$ and $q_\theta(z_{t}|h_{t-1}, a_{t-1}, o_{t})$ are multivariate normal distributions
over $z_t$ whose mean and diagonal variance are determined by neural networks.
For image data, the decoder $p_\theta(o_{t} \given z_{t}, a_{t-1})$
is a multivariate independent Bernoulli distribution whose
parameters are again determined by a neural network. For real-valued vectors we
use a normal distribution. 

When several inputs are passed to a neural network, they are concatenated to one vector.
ReLUs are used as nonlinearities between all layers.
Hidden layers are, if not otherwise stated, all of the same dimension as $h$.
Batch normalization was used between layers for experiments on Atari but not on
Mountain Hike as they significantly hurt performance.
All \glspl{RNN} are GRUs.

Encoding functions $\varphi^o$, $\varphi^a$ and $\varphi^z$ are used to encode single
observations, actions and latent states $z$ before they are passed into other networks.

To encode visual observations, we use the the same convolutional network as proposed by
\citet{mnih2015human}, but with only $32$ instead of $64$ channels in the final layer. The
transposed convolutional network of the decoder has the reversed structure. The
decoder is preceeded by an additional fully connected layer which outputs the
required dimension (1568 for Atari's $84 \times 84$ observations).

For observations in $\mathbb{R}^2$ we used two fully connected layers of size 64
as encoder. As decoder we used the same structure as for $p_\theta(z|\dots)$ and
$q_\phi(z|\dots)$ which are all three normal distributions: One joint fully
connected layer and two separated fully connected heads, one for the mean, one
for the variance. The output of the variance layer is passed through a softplus
layer to force positivity.

Actions are encoded using one fully connected layer of size 128 for Atari and
size 64 for Mountain Hike.
Lastly, $z$ is encoded before being passed into networks by one fully connected
layer of the same size as $h$.

The policy is one fully connected layer whose size is determined by the
actions space, i.e. up to 18 outputs with softmax for Atari and only 2 
outputs for the learned mean for Mountain Hike, together with a learned
variance. 
The value function is one fully connected layer of size 1.

A2C used $n_e=16$ parallel environments and $n_s=5$-step learning for a total batch size of 80.
Hyperparameters were tuned on \emph{Chopper Command}. The learning rate of both \gls{DVRL} and \gls{ADR-A2C} was independently tuned
on the set of values $\{3 \times 10^{-5}, 1 \times 10^{-4}, 2 \times 10^{-4}, 3 \times 10^{-4}, 6 \times 10^{-4}, 9 \times 10^{-4}\}$ with $2 \times 10^{-4}$ being
chosen for \gls{DVRL} on Atari and $1 \times 10^{-4}$ for \gls{DVRL} on
MountainHike and \gls{RNN} on both environments.
Without further tuning, we set $\lambda^H=0.01$ and $\lambda^V=0.5$ as is commonly used.

As optimizer we use RMSProp with $\alpha=0.99$. We clip gradients at a value of
$0.5$. The discount factor of the control problem is set to $\gamma=0.99$ and
lastly, we use 'orthogonal' initialization for the network weights.

The source code will be release in the future.
% The source code can be found here: \url{https://github.com/oxwhirl/Deep-Variational-Reinforcement-Learning/}

\subsection{Additional Experiments and Visualisations}

Table \ref{tab:ap:atari-det} shows the on \emph{deterministic} and flickering
Atari, averaged over 5 random seeds. The values for \gls{DRQN} and \gls{ADRQN}
are taken from the respective papers. Note that \gls{DRQN} and \gls{ADRQN} rely
on Q-learning instead of \gls{A2C}, so the results are not directly comparable.

\begin{figure*}
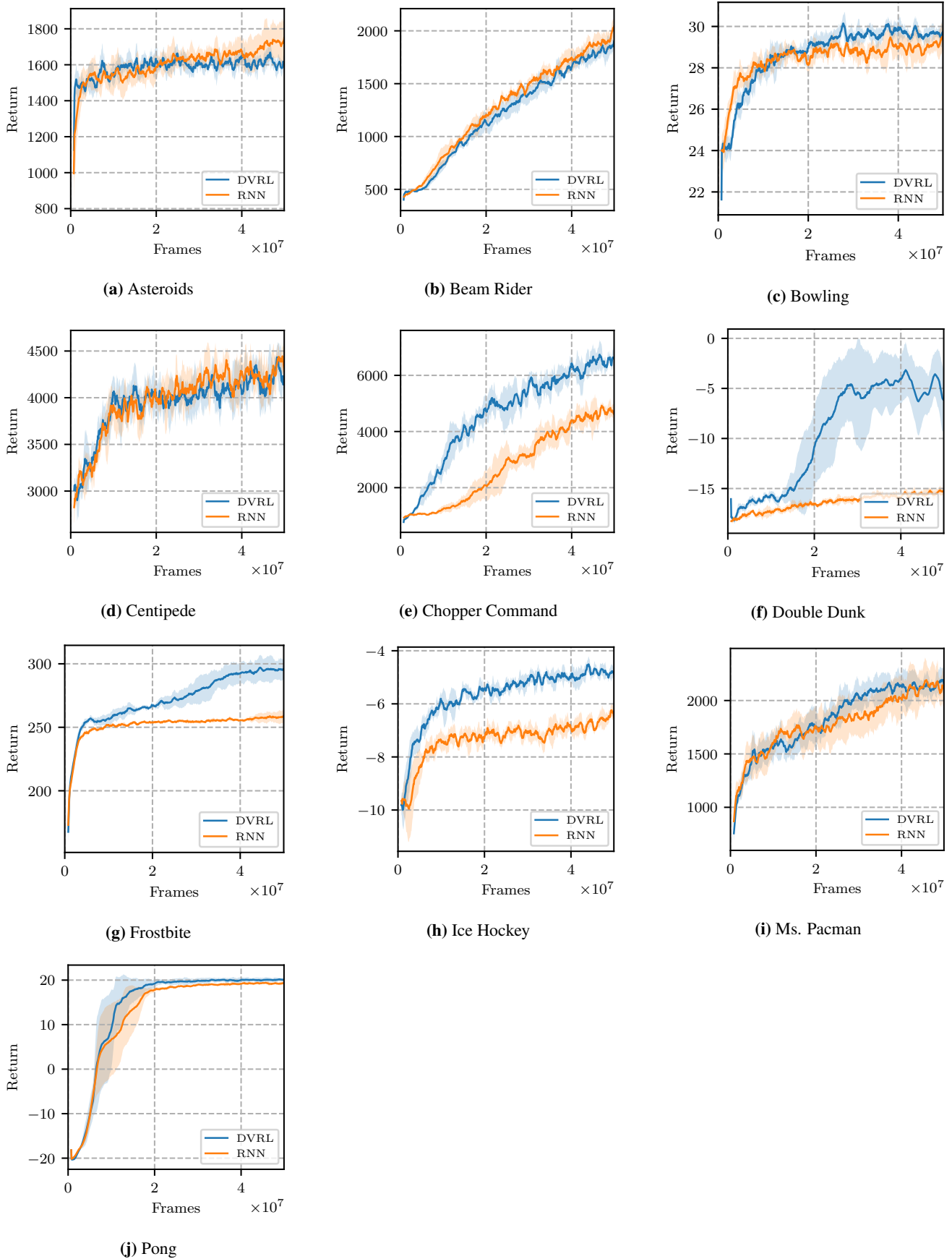

  \begin{subfigure}{6cm}
    \centering\includegraphics[width=0.95\textwidth]{{{images/Experiments/AsteroidsNoFrameskip-v4-result.true-}}}
    \caption{Asteroids}
  \end{subfigure}
  \begin{subfigure}{6cm}
    \centering\includegraphics[width=0.95\textwidth]{{{images/Experiments/BeamRiderNoFrameskip-v4-result.true-}}}
    \caption{Beam Rider}
  \end{subfigure}
  \begin{subfigure}{6cm}
    \centering\includegraphics[width=0.95\textwidth]{{{images/Experiments/BowlingNoFrameskip-v4-result.true-}}}
    \caption{Bowling}
  \end{subfigure}
  \begin{subfigure}{6cm}
    \centering\includegraphics[width=0.95\textwidth]{{{images/Experiments/CentipedeNoFrameskip-v4-result.true-}}}
    \caption{Centipede}
  \end{subfigure}
  \begin{subfigure}{6cm}
    \centering\includegraphics[width=0.95\textwidth]{{{images/Experiments/ChopperCommandNoFrameskip-v4-result.true-}}}
    \caption{Chopper Command}
  \end{subfigure}
  \begin{subfigure}{6cm}
    \centering\includegraphics[width=0.95\textwidth]{{{images/Experiments/DoubleDunkNoFrameskip-v4-result.true-}}}
    \caption{Double Dunk}
  \end{subfigure}
  \begin{subfigure}{6cm}
    \centering\includegraphics[width=0.95\textwidth]{{{images/Experiments/FrostbiteNoFrameskip-v4-result.true-}}}
    \caption{Frostbite}
  \end{subfigure}
  \begin{subfigure}{6cm}
    \centering\includegraphics[width=0.95\textwidth]{{{images/Experiments/IceHockeyNoFrameskip-v4-result.true-}}}
    \caption{Ice Hockey}
  \end{subfigure}
  \begin{subfigure}{6cm}
    \centering\includegraphics[width=0.95\textwidth]{{{images/Experiments/MsPacmanNoFrameskip-v4-result.true-}}}
    \caption{Ms. Pacman}
  \end{subfigure}
  \begin{subfigure}{6cm}
    \centering\includegraphics[width=0.95\textwidth]{{{images/Experiments/PongNoFrameskip-v4-result.true-}}}
    \caption{Pong}
  \end{subfigure}
  \caption{Training curves on the full set of evaluated Atari games, in the case
  of flickering and \emph{deterministic} environments.}
  \label{fig:ap:atari-det}
\end{figure*}

\begin{figure*}
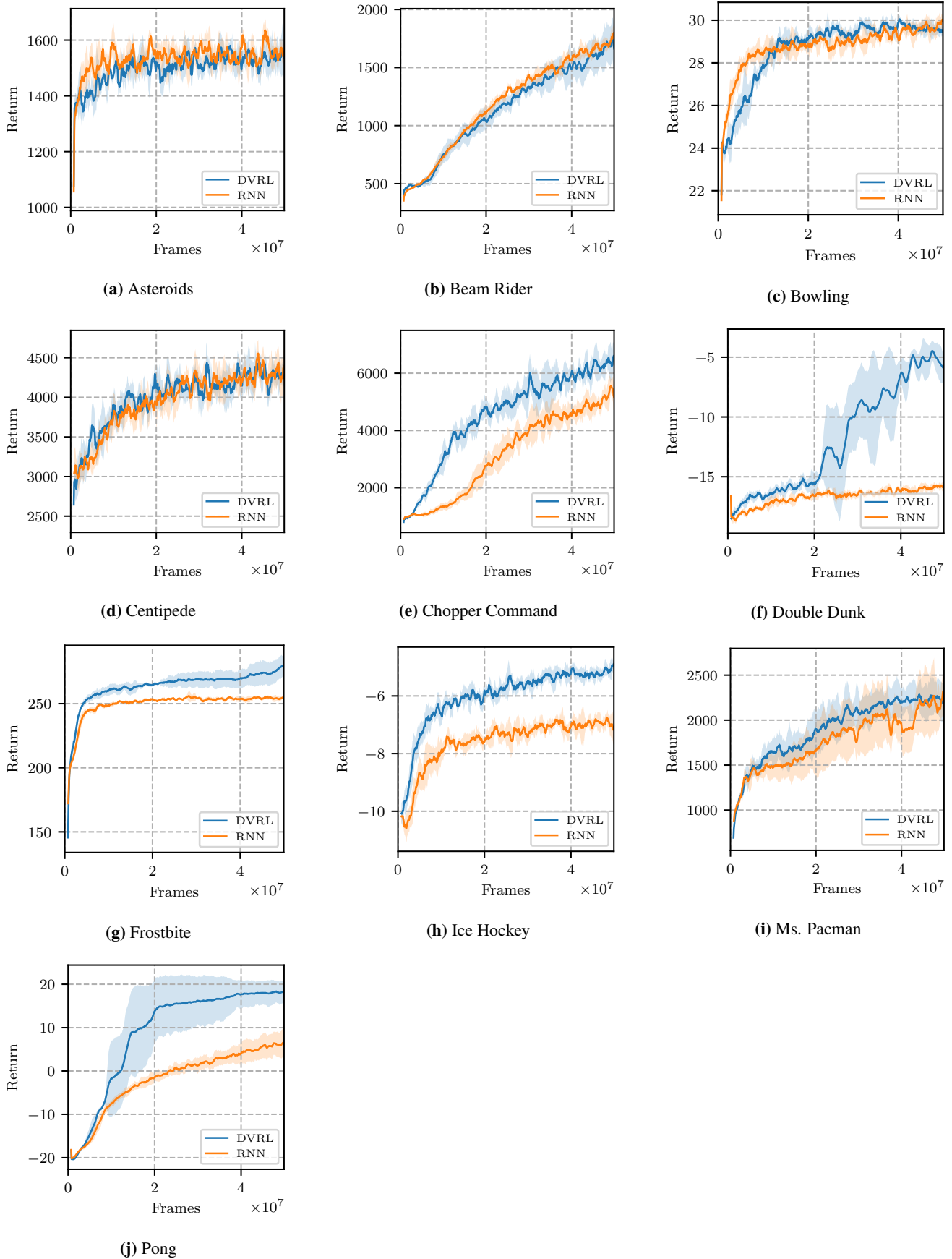

  \begin{subfigure}{6cm}
    \centering\includegraphics[width=0.95\textwidth]{{{images/Experiments/AsteroidsNoFrameskip-v0-result.true-}}}
    \caption{Asteroids}
  \end{subfigure}
  \begin{subfigure}{6cm}
    \centering\includegraphics[width=0.95\textwidth]{{{images/Experiments/BeamRiderNoFrameskip-v0-result.true-}}}
    \caption{Beam Rider}
  \end{subfigure}
  \begin{subfigure}{6cm}
    \centering\includegraphics[width=0.95\textwidth]{{{images/Experiments/BowlingNoFrameskip-v0-result.true-}}}
    \caption{Bowling}
  \end{subfigure}
  \begin{subfigure}{6cm}
    \centering\includegraphics[width=0.95\textwidth]{{{images/Experiments/CentipedeNoFrameskip-v0-result.true-}}}
    \caption{Centipede}
  \end{subfigure}
  \begin{subfigure}{6cm}
    \centering\includegraphics[width=0.95\textwidth]{{{images/Experiments/ChopperCommandNoFrameskip-v0-result.true-}}}
    \caption{Chopper Command}
  \end{subfigure}
  \begin{subfigure}{6cm}
    \centering\includegraphics[width=0.95\textwidth]{{{images/Experiments/DoubleDunkNoFrameskip-v0-result.true-}}}
    \caption{Double Dunk}
  \end{subfigure}
  \begin{subfigure}{6cm}
    \centering\includegraphics[width=0.95\textwidth]{{{images/Experiments/FrostbiteNoFrameskip-v0-result.true-}}}
    \caption{Frostbite}
  \end{subfigure}
  \begin{subfigure}{6cm}
    \centering\includegraphics[width=0.95\textwidth]{{{images/Experiments/IceHockeyNoFrameskip-v0-result.true-}}}
    \caption{Ice Hockey}
  \end{subfigure}
  \begin{subfigure}{6cm}
    \centering\includegraphics[width=0.95\textwidth]{{{images/Experiments/MsPacmanNoFrameskip-v0-result.true-}}}
    \caption{Ms. Pacman}
  \end{subfigure}
  \begin{subfigure}{6cm}
    \centering\includegraphics[width=0.95\textwidth]{{{images/Experiments/PongNoFrameskip-v0-result.true-}}}
    \caption{Pong}
  \end{subfigure}
  \caption{Training curves on the full set of evaluated Atari games, in the case
  of flickering and \emph{stochastic} environments.}
  \label{fig:ap:Atari-stoch}
\end{figure*}

Figure \ref{fig:ap:atari-det} and \ref{fig:ap:Atari-stoch} show individual
learning curves for all 10 Atari games, either for the deterministic or the
stochastic version of the games.

\begin{table}[!htb]
  \centering
  \caption{Final results on deterministic and flickering Atari environments,
    averaged over 5 random seeds. Bold numbers indicate statistical significance
  at the 5\% level when comparing \gls{DVRL} and \gls{RNN}. The values for
  \gls{DRQN} and \gls{ADRQN} are taken from the respective papers.} 
  \label{tab:ap:atari-det}
  \begin{tabular}{lll|ll}
    \toprule
    Env       & \gls{DVRL}$(\pm std)$                                                   & \gls{RNN}                                   & \gls{DRQN}        & \gls{ADRQN}      \\
    \midrule
    Pong      & $\bm{20.07} (\pm 0.39)$                                                 & $19.3 (\pm 0.26)$                           & $12.1 (\pm 2.2)$  & $7 (\pm 4.6)$    \\
    Chopper   & $\bm{6619} (\pm 532)$                                                   & $4619(\pm 306)$                             & $1330 (\pm 294)$  & $1608 (\pm 707)$ \\
    MsPacman  & $2156 (\pm 127)$                                                        & $2113(\pm 135)$                             & $1739 (\pm 942)$  &                  \\
    Centipede & $4171 (\pm 127)$                                                        & $4283 (\pm187)$                             & $4319 (\pm 4378)$ &                  \\
    BeamRider & $1901(\pm 67)$                                                          & $\bm{2041} (\pm 81)$                        & $618 (\pm 115)$   &                  \\
    Frostbite & $\bm{296} (\pm 8.2)$                                                    & $259 (\pm 5.7)$                             & $414 (\pm 494)$   & $2002 (\pm 734)$ \\
    Bowling   & $29.74 (\pm 0.49)$                                                      & $29.38 (\pm 0.52)$                          & $65 (\pm 13)$     &                  \\
    IceHockey & $\bm{-4.87} (\pm 0.24)$                                                 & $ -6.49 (\pm 0.27)$                         & $-5.4 (\pm 2.7)$  &                  \\
    DDunk     & $\bm{-6.08} (\pm 3.08)$                                                 & $ -15.25 (\pm 0.51)$                        & $-14 (\pm 2.5)$   & $-13 (\pm 3.6)$  \\
    Asteroids & $1610 (\pm 63)$                                                         & $\bm{1750} (\pm 97)$                       & $1032 (\pm 410)$  & $1040 (\pm 431)$ \\
    \bottomrule
  \end{tabular}
\end{table}

\subsection{Computational Speed}
The approximate training speed in frames per second (FPS) is on one GPU on a
dgx1 for Atari:
\begin{itemize}
\itemsep0em 
\item \gls{RNN}: 124k FPS
\item \gls{DVRL} (1 Particle): 64k FPS
\item \gls{DVRL} (10 Particles): 48k FPS
\item \gls{DVRL} (30 Particle): 32k FPS
\end{itemize}

\subsection{Model Predictions}

\begin{figure*}
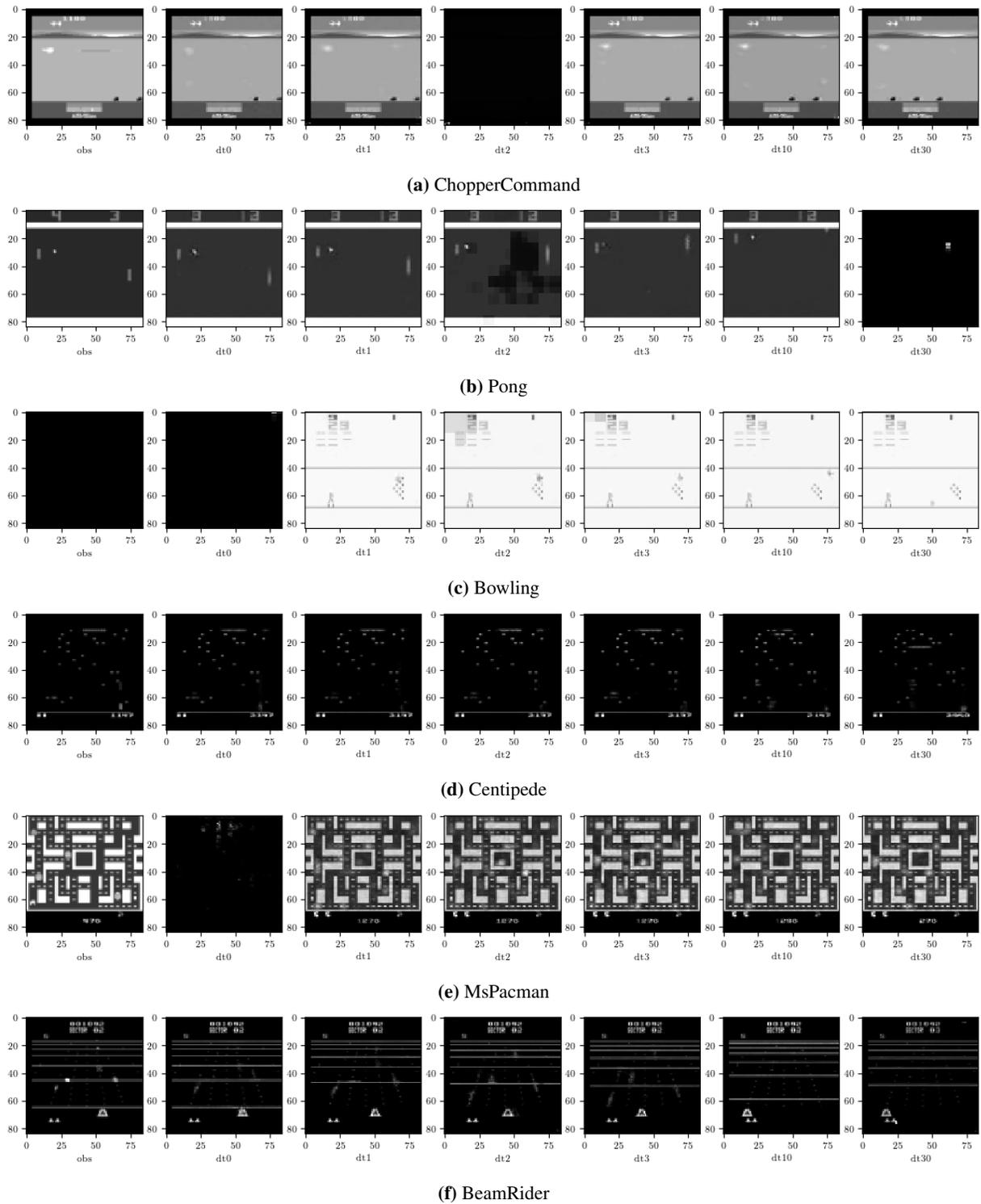


  \begin{subfigure}{\textwidth}
    \centering\includegraphics[width=0.95\textwidth]{{{images/ModelPredictions/ChCom1}}}
    \caption{ChopperCommand}
  \end{subfigure}

  \begin{subfigure}{\textwidth}
    \centering\includegraphics[width=0.95\textwidth]{{{images/ModelPredictions/Pong0}}}
    \caption{Pong}
  \end{subfigure}

  % \begin{subfigure}{\textwidth}
  %   \centering\includegraphics[width=0.95\textwidth]{{{images/ModelPredictions/BeamRider0}}}
  %   \caption{BeamRider}
  % \end{subfigure}

  % \begin{subfigure}{\textwidth}
  %   \centering\includegraphics[width=0.95\textwidth]{{{images/ModelPredictions/Bowling0}}}
  % \end{subfigure}
  \begin{subfigure}{\textwidth}
    \centering\includegraphics[width=0.95\textwidth]{{{images/ModelPredictions/Bowling1}}}
    \caption{Bowling}
  \end{subfigure}

  \begin{subfigure}{\textwidth}
    \centering\includegraphics[width=0.95\textwidth]{{{images/ModelPredictions/Centipede0}}}
    \caption{Centipede}
  \end{subfigure}
  % \begin{subfigure}{\textwidth}
  %   \centering\includegraphics[width=0.95\textwidth]{{{images/ModelPredictions/Centipede1}}}
  %   \caption{Centipede}
  % \end{subfigure}

  % \begin{subfigure}{\textwidth}
  %   \centering\includegraphics[width=0.95\textwidth]{{{images/ModelPredictions/ChCom0}}}
  % \end{subfigure}

  \begin{subfigure}{\textwidth}
    \centering\includegraphics[width=0.95\textwidth]{{{images/ModelPredictions/MsPacman1}}}
    \caption{MsPacman}
  \end{subfigure}

  \begin{subfigure}{\textwidth}
    \centering\includegraphics[width=0.95\textwidth]{{{images/ModelPredictions/BeamRider1}}}
    \caption{BeamRider}
  \end{subfigure}

  \caption{Reconstructions and predictions using the learned generative model
    for several Atari games. First column: Current obseration (potentially blank). Second column:
    Encoded and decoded reconstruction of the current observation. Columns 3 to
    7: Predicted observations using the learned generative model for timesteps
    $dt\in\{0,1,2,3,10,30\}$ into the future.}
  \label{fig:ap:model-predictions}
\end{figure*}

In Figure \ref{fig:ap:model-predictions} we show reconstructed and predicted
images from the \gls{DVRL} model for several Atari games. The current
observation is in the leftmost column. The second column ('dt0') shows the
reconstruction after encoding and decoding the current observation. For the
further columns, we make use of the learned generative model to predict future
observations. For simplicity we repeat the last action. Columns 2 to 7 show
predicted observations for $dt\in\{1,2,3,10,30\}$ unrolled timesteps.
The model was trained as explained in Section \ref{sec:exp:atari}. The
reconstructed and predicted images are a weighted average over all 16 particles.

Note that the model is able to correctly predict features of future
observations, for example the movement of the cars in ChopperCommand, the
(approximate) ball position in Pong or the missing pins in Bowling.
Furthermore, it is able to do so, even if the current observation is blank like
in Bowling.
The model has also correctly learned to randomly predict blank observations.

It can remember feature of the current state fairly well, like the
positions of barriers (white dots) in Centipede. On the other hand, it clearly
struggles with the amount of information present in MsPacman like the positions
of all previously eaten fruits or the location of the ghosts. 

% \begin{figure*}
%   \begin{subfigure}{\textwidth}

%     \centering\includegraphics[width=0.95\textwidth]{{{images/ModelPredictions/Frostbite0}}}
%   \end{subfigure}
%   \begin{subfigure}{\textwidth}
%     \centering\includegraphics[width=0.95\textwidth]{{{images/ModelPredictions/Frostbite1}}}
%     \caption{Frostbite}
%   \end{subfigure}

%   \begin{subfigure}{\textwidth}
%     \centering\includegraphics[width=0.95\textwidth]{{{images/ModelPredictions/IceHockey0}}}
%   \end{subfigure}
%   \begin{subfigure}{\textwidth}
%     \centering\includegraphics[width=0.95\textwidth]{{{images/ModelPredictions/IceHockey1}}}
%     \caption{IceHockey}
%   \end{subfigure}

%   \begin{subfigure}{\textwidth}
%     \centering\includegraphics[width=0.95\textwidth]{{{images/ModelPredictions/MsPacman0}}}
%   \end{subfigure}
%   \begin{subfigure}{\textwidth}
%     \centering\includegraphics[width=0.95\textwidth]{{{images/ModelPredictions/MsPacman1}}}
%     \caption{MsPacman}
%   \end{subfigure}

%   \begin{subfigure}{\textwidth}
%     \centering\includegraphics[width=0.95\textwidth]{{{images/ModelPredictions/Pong0}}}
%     \caption{Pong}
%   \end{subfigure}
%   \begin{subfigure}{\textwidth}
%     \centering\includegraphics[width=0.95\textwidth]{{{images/ModelPredictions/Pong1}}}
%     \caption{Pong}
%   \end{subfigure}

%   \caption{Training curves on the full set of evaluated Atari games, in the case
%   of flickering and \emph{stochastic} environments.}
%   % \label{fig:ap:Atari-stoch}
% \end{figure*}

%% file: core/8_appendix_algorithms.tex
\section{Algorithms}
\label{appendix:alg}

Algorithm \ref{alg:enc_dvp} details the recurrent (belief) state computation (i.e. history
encoder) for \gls{DVRL}. Algorithm \ref{alg:enc_rnn} details the recurrent state
computation for \gls{RNN}. 
Algorithm \ref{alg:training} describes the overall training algorithm that
either uses one or the other to aggregate the history. Despite looking
complicated, it is just a very detailed implementation of $n$-step A2C with the
additional changes: Inclusion of $\mathcal{L}^{\text{ELBO}}$ and inclusing of the
option to not delete the computation graph to allow longer backprop in $n$-step
A2C. 

Results for also using the reconstruction loss $\mathcal{L}^{\text{ENC}}$ for
the \gls{RNN} based encoder aren't shown in the paper as they reliably performed
worse than \gls{RNN} without reconstruction loss.

\begin{algorithm}[ht!]
  \caption{\gls{DVRL} encoder }
  \begin{algorithmic}
    \STATE {\bfseries Input:} Previous state $\hat{b}_{t-1}$, observation $o_t$,
    action $a_{t-1}$
    \STATE Unpack $w^{1:K}_{t-1}, z^{1:K}_{t-1}, h_{t-1}^{1:K}, \hat{h}_{t-1} \leftarrow \hat{b}_{t-1}$
    \STATE $x^o \leftarrow \varphi^o_\theta(o_t)$
    \STATE $x^a \leftarrow \varphi^a_\theta(a_{t-1})$

    \FOR{$k=1$ {\bfseries to} $K$}
    \STATE Sample $h_{t-1}^k\sim h_{t-1}^{1:K}$ based on weights
    \STATE Sample $z_t^k\sim q_\theta(z^k_t|h^k_{t-1},x^o,x^a)$
    \STATE $x^z \leftarrow \varphi^z_\theta(z_t)$ 
    \STATE $w^k_j \leftarrow p_\theta(z^k_t|h^k_{t-1},x^a) p_\theta(o_t|h^k_t, x^z, x^a) / q_\theta(z^k_t|h^k_{t-1},x^o,x^a)$
    \STATE $h_t^k \leftarrow \text{GRU}(h_{t-1}^k, x^z, x^o, x^a)$
    \ENDFOR

    \STATE $\mathcal{L}^{\text{ELBO}}_t \leftarrow - \log \sum_k w^k_t - \log(K)$
    \STATE $\hat{h}_t \leftarrow \text{GRU}(\text{Concat}(w_t^k,x^z,h_t^k)_{k=1}^K \text{passed sequentially})$
    \STATE Pack $\hat{b}_t \leftarrow w^{1:K}_{t}, z^{1:K}_{t}, h_t^{1:K}, \hat{h}_t$
    \STATE \COMMENT{When $V$ or $\pi$ is conditioned on $\hat{b}_t$, the summary
    $\hat{h}_t$ is used.}
   
    \STATE {\bfseries Output:} $\hat{b}_t, \mathcal{L}^{\text{ELBO}}_t$
  \end{algorithmic}
  \label{alg:enc_dvp}
\end{algorithm}

\begin{algorithm}[tb]
  \caption{\gls{RNN} encoder}
  \label{alg:enc_rnn}
  \begin{algorithmic}
    \STATE {\bfseries Input:} Previous state $h_{j-1}$, observation
    $o_j$, action $a_{j-1}$
    \STATE $x^o \leftarrow \varphi^o_\theta(o_t)$
    \STATE $x^a \leftarrow \varphi^a_\theta(a_{t-1})$
    \STATE $\hat{b}_j \leftarrow \text{GRU}_\theta(\hat{b}_{j-1}, x^o, x^a)$
    \STATE $\mathcal{L}^{\text{ENC}}_j \leftarrow - \log
    p_\theta(o_j|\hat{b}_{j-1})$
    \STATE {\bfseries Output:} $h_j, \mathcal{L}^{\text{ENC}}_j$
  \end{algorithmic}
\end{algorithm}

\begin{algorithm}[tb]
  \caption{Training Algorithm}
  \label{alg:training}
  \begin{algorithmic}
    \STATE {\bfseries Input:} Environment $\mathrm{Env}$, Encoder
    $\mathrm{Enc}_{\theta,\phi}$ (either RNN or \gls{DVRL})
    \STATE Initialize observation $o_1$ from $\mathrm{Env}$.
    \STATE Initialize encoder latent state $s_0\leftarrow s_0^{init}$ as either $h_0$ (for \gls{RNN}) or
    $\hat{b}_{0,\theta}$ (for \gls{DVRL}) 
    \STATE Initialize action $a_0=0$ to no-op
    \STATE Set $s_0', a_0' \leftarrow s_0, a_0$. 
    \STATE \COMMENT{The distinction between $s'_t$ and $s_t$ is necessary when the
      environment resets at time $t$.}
    \REPEAT
    \STATE $\mathcal{L}^{Enc}_j, a_j, a_j', s_j, s_j', o_{j+1}, r_{j+1},
    done_{j+1}\leftarrow NULL \quad j=1\dots n$
    \STATE \COMMENT{Run $n$ steps forward:}
    \FOR{$j=1$ {\bfseries to} $n$}
    \STATE $s_j, \mathcal{L}^{\text{ELBO}}_j \leftarrow \mathrm{Enc}_{\theta,\phi}(o_{j}, a_{j-1}', s'_{j-1})$
    \STATE Sample $a_j \sim \pi_\rho(a_j|s_j)$
    \STATE $o_{j+1}, r_{j+1}, done_{j+1} \leftarrow \text{Env}(a_j)$

    \IF{$done_{j+1}$}
    \STATE $s'_j, a_j' \leftarrow s_0^{init}, 0$
    \STATE \COMMENT{$s_j$ is still available to compute $V_\eta(s_j)$}
    \STATE $o_{j+1}\leftarrow$ Reset $\mathrm{Env}()$
    \ELSE
    \STATE $s'_j, a'_j \leftarrow s_j, a_j$
    \ENDIF
    \ENDFOR

    \STATE \COMMENT{Compute targets}
    \STATE $s_{n+1} \leftarrow \mathrm{Enc}_{\theta,\phi}(o_{n+1}, a_{n}',
    s'_{n})$
    \STATE $Q^{target}_{n+1} \leftarrow V_\eta(s_{n+1}).detach()$

    \FOR{$j=n$ {\bfseries to} $1$}
    \STATE $Q^{target}_j \leftarrow \gamma \cdot Q^{target}_{j+1}$
    \IF{$done_{j+1}$}
    \STATE $Q^{target}_j \leftarrow 0$
    \ENDIF
    \STATE $Q^{target}_j \leftarrow Q^{target}_j + r_{j+1}$
    \ENDFOR

    \STATE \COMMENT{Compute losses}

    \FOR{$j=n$ {\bfseries to} $1$}
    \STATE $\mathcal{L}^V_j \leftarrow (Q^{target}_j-V_\eta(s_{j}))^2$
    \STATE $\mathcal{L}^A_j \leftarrow - \log\pi_\rho(a_j|s_{j})(Q^{target}_j-V_\eta(s_{j}))$
    \STATE $\mathcal{L}^H_j \leftarrow - \text{Entropy}(\pi_\rho(\cdot|s_j))$
    \ENDFOR
    \STATE $\mathcal{J} \leftarrow \sum_j (\lambda^V \mathcal{L}^V_j + \mathcal{L}^A_j +
    \lambda^H\mathcal{L}^H_j + \lambda^E \mathcal{L}^{\text{ELBO}}_j$)
    \STATE $\text{TakeGradientStep}(\mathcal{\nabla J})$
    \STATE Delete or save computation graph of $s_n$ to determine backpropagation length
    \STATE $a_0', s_0' \leftarrow a_{n}, s_{n}$
    \STATE $o_1 \leftarrow o_{n+1}$
    \UNTIL converged
  \end{algorithmic}
\end{algorithm}

%% file: main.bbl
\ifdefined\DeclarePrefChars\DeclarePrefChars{'’-}\else\fi
\begin{thebibliography}{50}
\providecommand{\natexlab}[1]{#1}
\providecommand{\url}[1]{\texttt{#1}}
\expandafter\ifx\csname urlstyle\endcsname\relax
  \providecommand{\doi}[1]{doi: #1}\else
  \providecommand{\doi}{doi: \begingroup \urlstyle{rm}\Url}\fi

\bibitem[Astrom(1965)]{astrom1965optimal}
Astrom, Karl~J.
\newblock Optimal control of markov decision processes with incomplete state
  estimation.
\newblock \emph{Journal of mathematical analysis and applications},
  10:\penalty0 174--205, 1965.

\bibitem[Azizzadenesheli et~al.(2016)Azizzadenesheli, Lazaric, and
  Anandkumar]{azizzadenesheli2016reinforcement}
Azizzadenesheli, Kamyar, Lazaric, Alessandro, and Anandkumar, Animashree.
\newblock Reinforcement learning of pomdps using spectral methods.
\newblock \emph{arXiv preprint 1602.07764}, 2016.

\bibitem[Babayan et~al.(2018)Babayan, Uchida, and Gershman]{babayan2018belief}
Babayan, Benedicte~M, Uchida, Naoshige, and Gershman, Samuel~J.
\newblock Belief state representation in the dopamine system.
\newblock \emph{Nature communications}, 9\penalty0 (1):\penalty0 1891, 2018.

\bibitem[Bakker(2002)]{bakker2002reinforcement}
Bakker, Bram.
\newblock Reinforcement learning with long short-term memory.
\newblock In \emph{Advances in neural information processing systems}, pp.\
  1475--1482, 2002.

\bibitem[Barto et~al.(1995)Barto, Bradtke, and Singh]{barto1995learning}
Barto, Andrew~G, Bradtke, Steven~J, and Singh, Satinder~P.
\newblock Learning to act using real-time dynamic programming.
\newblock \emph{Artificial intelligence}, 72\penalty0 (1-2):\penalty0 81--138,
  1995.

\bibitem[Bellemare et~al.(2014)Bellemare, Veness, and
  Talvitie]{bellemare2014skip}
Bellemare, Marc, Veness, Joel, and Talvitie, Erik.
\newblock Skip context tree switching.
\newblock In \emph{International Conference on Machine Learning}, pp.\
  1458--1466, 2014.

\bibitem[Bellemare(2015)]{bellemare2015count}
Bellemare, Marc~G.
\newblock Count-based frequency estimation with bounded memory.
\newblock In \emph{Twenty-Fourth International Joint Conference on Artificial
  Intelligence}, 2015.

\bibitem[Bellemare et~al.(2013)Bellemare, Naddaf, Veness, and
  Bowling]{bellemare2013arcade}
Bellemare, Marc~G, Naddaf, Yavar, Veness, Joel, and Bowling, Michael.
\newblock The arcade learning environment: An evaluation platform for general
  agents.
\newblock \emph{Journal of Artificial Intelligence Research}, 47:\penalty0
  253--279, 2013.

\bibitem[Burda et~al.(2016)Burda, Grosse, and
  Salakhutdinov]{burda2016importance}
Burda, Yuri, Grosse, Roger, and Salakhutdinov, Ruslan.
\newblock Importance weighted autoencoders.
\newblock In \emph{ICLR}, 2016.

\bibitem[Chung et~al.(2015)Chung, Kastner, Dinh, Goel, Courville, and
  Bengio]{chung2015recurrent}
Chung, Junyoung, Kastner, Kyle, Dinh, Laurent, Goel, Kratarth, Courville,
  Aaron~C, and Bengio, Yoshua.
\newblock A recurrent latent variable model for sequential data.
\newblock In \emph{Advances in neural information processing systems}, 2015.

\bibitem[Coquelin et~al.(2009)Coquelin, Deguest, and
  Munos]{coquelin2009particle}
Coquelin, Pierre-Arnaud, Deguest, Romain, and Munos, R{\'e}mi.
\newblock Particle filter-based policy gradient in pomdps.
\newblock In \emph{NIPS}, 2009.

\bibitem[Deisenroth \& Peters(2012)Deisenroth and
  Peters]{deisenroth2012solving}
Deisenroth, Marc~Peter and Peters, Jan.
\newblock Solving nonlinear continuous state-action-observation pomdps for
  mechanical systems with gaussian noise.
\newblock 2012.

\bibitem[Dhariwal et~al.(2017)Dhariwal, Hesse, Klimov, Nichol, Plappert,
  Radford, Schulman, Sidor, and Wu]{dhariwal2017openaibaselines}
Dhariwal, Prafulla, Hesse, Christopher, Klimov, Oleg, Nichol, Alex, Plappert,
  Matthias, Radford, Alec, Schulman, John, Sidor, Szymon, and Wu, Yuhuai.
\newblock Openai baselines, 2017.

\bibitem[Doshi-Velez et~al.(2015)Doshi-Velez, Pfau, Wood, and
  Roy]{doshi2015bayesian}
Doshi-Velez, Finale, Pfau, David, Wood, Frank, and Roy, Nicholas.
\newblock Bayesian nonparametric methods for partially-observable reinforcement
  learning.
\newblock \emph{IEEE transactions on pattern analysis and machine
  intelligence}, 37\penalty0 (2):\penalty0 394--407, 2015.

\bibitem[Doucet \& Johansen(2009)Doucet and Johansen]{doucet2009tutorial}
Doucet, Arnaud and Johansen, Adam~M.
\newblock A tutorial on particle filtering and smoothing: Fifteen years later.
\newblock \emph{Handbook of nonlinear filtering}, 12\penalty0
  (656-704):\penalty0 3, 2009.

\bibitem[Foerster et~al.(2016)Foerster, Assael, de~Freitas, and
  Whiteson]{foerster2016learning}
Foerster, Jakob~N, Assael, Yannis~M, de~Freitas, Nando, and Whiteson, Shimon.
\newblock Learning to communicate to solve riddles with deep distributed
  recurrent q-networks.
\newblock \emph{arXiv preprint 1602.02672}, 2016.

\bibitem[Hausknecht \& Stone(2015)Hausknecht and Stone]{hausknecht2015deep}
Hausknecht, Matthew and Stone, Peter.
\newblock Deep recurrent q-learning for partially observable {MDP}s.
\newblock In \emph{2015 AAAI Fall Symposium Series}, 2015.

\bibitem[Heess et~al.(2015)Heess, Hunt, Lillicrap, and Silver]{heess2015memory}
Heess, Nicolas, Hunt, Jonathan~J, Lillicrap, Timothy~P, and Silver, David.
\newblock Memory-based control with recurrent neural networks.
\newblock \emph{arXiv preprint 1512.04455}, 2015.

\bibitem[Jaderberg et~al.(2016)Jaderberg, Mnih, Czarnecki, Schaul, Leibo,
  Silver, and Kavukcuoglu]{jaderberg2016reinforcement}
Jaderberg, Max, Mnih, Volodymyr, Czarnecki, Wojciech~Marian, Schaul, Tom,
  Leibo, Joel~Z, Silver, David, and Kavukcuoglu, Koray.
\newblock Reinforcement learning with unsupervised auxiliary tasks.
\newblock \emph{arXiv preprint 1611.05397}, 2016.

\bibitem[Kaelbling et~al.(1998)Kaelbling, Littman, and
  Cassandra]{kaelbling1998planning}
Kaelbling, Leslie~Pack, Littman, Michael~L, and Cassandra, Anthony~R.
\newblock Planning and acting in partially observable stochastic domains.
\newblock \emph{Artificial intelligence}, 101\penalty0 (1), 1998.

\bibitem[Karkus et~al.(2017)Karkus, Hsu, and Lee]{karkus2017qmdp}
Karkus, Peter, Hsu, David, and Lee, Wee~Sun.
\newblock Qmdp-net: Deep learning for planning under partial observability.
\newblock In \emph{Advances in Neural Information Processing Systems}, pp.\
  4697--4707, 2017.

\bibitem[Katt et~al.(2017)Katt, Oliehoek, and Amato]{katt2017learning}
Katt, Sammie, Oliehoek, Frans~A, and Amato, Christopher.
\newblock Learning in pomdps with monte carlo tree search.
\newblock In \emph{International Conference on Machine Learning}, 2017.

\bibitem[Kingma \& Welling(2014)Kingma and Welling]{kingma2014auto}
Kingma, Diederik~P and Welling, Max.
\newblock Auto-encoding variational {B}ayes.
\newblock In \emph{ICLR}, 2014.

\bibitem[Lample \& Chaplot(2017)Lample and Chaplot]{lample2017playing}
Lample, Guillaume and Chaplot, Devendra~Singh.
\newblock Playing fps games with deep reinforcement learning.
\newblock In \emph{AAAI}, pp.\  2140--2146, 2017.

\bibitem[Le et~al.(2018)Le, Igl, Jin, Rainforth, and Wood]{le2018autoencoding}
Le, Tuan~Anh, Igl, Maximilian, Jin, Tom, Rainforth, Tom, and Wood, Frank.
\newblock Auto-encoding sequential {M}onte {C}arlo.
\newblock In \emph{ICLR}, 2018.

\bibitem[Maddison et~al.(2017)Maddison, Lawson, Tucker, Heess, Norouzi, Mnih,
  Doucet, and Teh]{maddison2017filtering}
Maddison, Chris~J, Lawson, John, Tucker, George, Heess, Nicolas, Norouzi,
  Mohammad, Mnih, Andriy, Doucet, Arnaud, and Teh, Yee.
\newblock Filtering variational objectives.
\newblock In \emph{Advances in Neural Information Processing Systems}, 2017.

\bibitem[McAllester \& Singh(1999)McAllester and
  Singh]{mcallester1999approximate}
McAllester, David~A and Singh, Satinder.
\newblock Approximate planning for factored pomdps using belief state
  simplification.
\newblock In \emph{Proceedings of the Fifteenth conference on Uncertainty in
  artificial intelligence}, 1999.

\bibitem[McCallum \& Ballard(1996)McCallum and
  Ballard]{mccallum1996reinforcement}
McCallum, Andrew~Kachites and Ballard, Dana.
\newblock \emph{Reinforcement learning with selective perception and hidden
  state}.
\newblock PhD thesis, University of Rochester. Dept. of Computer Science, 1996.

\bibitem[McCallum(1993)]{mccallum1993overcoming}
McCallum, R~Andrew.
\newblock Overcoming incomplete perception with utile distinction memory.
\newblock In \emph{Proceedings of the Tenth International Conference on Machine
  Learning}, pp.\  190--196, 1993.

\bibitem[Messias \& Whiteson(2017)Messias and Whiteson]{messias2017dynamic}
Messias, Jo{\~a}o~V and Whiteson, Shimon.
\newblock Dynamic-depth context tree weighting.
\newblock In \emph{Advances in Neural Information Processing Systems}, pp.\
  3330--3339, 2017.

\bibitem[Mnih et~al.(2015)Mnih, Kavukcuoglu, Silver, Rusu, Veness, Bellemare,
  Graves, Riedmiller, Fidjeland, Ostrovski, et~al.]{mnih2015human}
Mnih, Volodymyr, Kavukcuoglu, Koray, Silver, David, Rusu, Andrei~A, Veness,
  Joel, Bellemare, Marc~G, Graves, Alex, Riedmiller, Martin, Fidjeland,
  Andreas~K, Ostrovski, Georg, et~al.
\newblock Human-level control through deep reinforcement learning.
\newblock \emph{Nature}, 518\penalty0 (7540):\penalty0 529, 2015.

\bibitem[Mnih et~al.(2016)Mnih, Badia, Mirza, Graves, Lillicrap, Harley,
  Silver, and Kavukcuoglu]{mnih2016asynchronous}
Mnih, Volodymyr, Badia, Adria~Puigdomenech, Mirza, Mehdi, Graves, Alex,
  Lillicrap, Timothy, Harley, Tim, Silver, David, and Kavukcuoglu, Koray.
\newblock Asynchronous methods for deep reinforcement learning.
\newblock In \emph{International Conference on Machine Learning}, 2016.

\bibitem[Naesseth et~al.(2018)Naesseth, Linderman, Ranganath, and
  Blei]{naesseth2017variational}
Naesseth, Christian~A, Linderman, Scott~W, Ranganath, Rajesh, and Blei,
  David~M.
\newblock Variational sequential monte carlo.
\newblock In \emph{AISTATS (To Appear)}, 2018.

\bibitem[Narasimhan et~al.(2015)Narasimhan, Kulkarni, and
  Barzilay]{narasimhan2015language}
Narasimhan, Karthik, Kulkarni, Tejas, and Barzilay, Regina.
\newblock Language understanding for text-based games using deep reinforcement
  learning.
\newblock \emph{arXiv preprint 1506.08941}, 2015.

\bibitem[Oliehoek et~al.(2008)Oliehoek, Spaan, Whiteson, and
  Vlassis]{oliehoek2008exploiting}
Oliehoek, Frans~A, Spaan, Matthijs~TJ, Whiteson, Shimon, and Vlassis, Nikos.
\newblock Exploiting locality of interaction in factored dec-pomdps.
\newblock In \emph{Proceedings of the 7th international joint conference on
  Autonomous agents and multiagent systems-Volume 1}, 2008.

\bibitem[Pineau et~al.(2003)Pineau, Gordon, Thrun, et~al.]{pineau2003point}
Pineau, Joelle, Gordon, Geoff, Thrun, Sebastian, et~al.
\newblock Point-based value iteration: An anytime algorithm for pomdps.
\newblock In \emph{IJCAI}, volume~3, 2003.

\bibitem[Rezende et~al.(2014)Rezende, Mohamed, and
  Wierstra]{rezende2014stochastic}
Rezende, Danilo~Jimenez, Mohamed, Shakir, and Wierstra, Daan.
\newblock Stochastic backpropagation and approximate inference in deep
  generative models.
\newblock In \emph{ICML}, 2014.

\bibitem[Roijers et~al.(2015)Roijers, Whiteson, and Oliehoek]{roijers2015point}
Roijers, Diederik~Marijn, Whiteson, Shimon, and Oliehoek, Frans~A.
\newblock Point-based planning for multi-objective pomdps.
\newblock In \emph{IJCAI}, pp.\  1666--1672, 2015.

\bibitem[Ross et~al.(2008)Ross, Pineau, Paquet, and Chaib-Draa]{ross2008online}
Ross, St{\'e}phane, Pineau, Joelle, Paquet, S{\'e}bastien, and Chaib-Draa,
  Brahim.
\newblock Online planning algorithms for pomdps.
\newblock \emph{Journal of Artificial Intelligence Research}, 32:\penalty0
  663--704, 2008.

\bibitem[Ross et~al.(2011)Ross, Pineau, Chaib-draa, and
  Kreitmann]{ross2011bayesian}
Ross, St{\'e}phane, Pineau, Joelle, Chaib-draa, Brahim, and Kreitmann, Pierre.
\newblock A bayesian approach for learning and planning in partially observable
  markov decision processes.
\newblock \emph{Journal of Machine Learning Research}, 2011.

\bibitem[Shani et~al.(2005)Shani, Brafman, and Shimony]{shani2005model}
Shani, Guy, Brafman, Ronen~I, and Shimony, Solomon~E.
\newblock Model-based online learning of pomdps.
\newblock In \emph{European Conference on Machine Learning}, pp.\  353--364.
  Springer, 2005.

\bibitem[Silver \& Veness(2010)Silver and Veness]{silver2010monte}
Silver, David and Veness, Joel.
\newblock Monte-carlo planning in large pomdps.
\newblock In \emph{Advances in neural information processing systems}, pp.\
  2164--2172, 2010.

\bibitem[Sohn et~al.(2015)Sohn, Lee, and Yan]{sohn2015learning}
Sohn, Kihyuk, Lee, Honglak, and Yan, Xinchen.
\newblock Learning structured output representation using deep conditional
  generative models.
\newblock In \emph{Advances in Neural Information Processing Systems}, pp.\
  3483--3491, 2015.

\bibitem[Tamar et~al.(2016)Tamar, Wu, Thomas, Levine, and
  Abbeel]{tamar2016value}
Tamar, Aviv, Wu, Yi, Thomas, Garrett, Levine, Sergey, and Abbeel, Pieter.
\newblock Value iteration networks.
\newblock In \emph{Advances in Neural Information Processing Systems}, pp.\
  2154--2162, 2016.

\bibitem[Thrun(2000)]{thrun2000monte}
Thrun, Sebastian.
\newblock Monte carlo pomdps.
\newblock In \emph{Advances in neural information processing systems}, pp.\
  1064--1070, 2000.

\bibitem[Wierstra et~al.(2007)Wierstra, Foerster, Peters, and
  Schmidhuber]{wierstra2007solving}
Wierstra, Daan, Foerster, Alexander, Peters, Jan, and Schmidhuber, Juergen.
\newblock Solving deep memory pomdps with recurrent policy gradients.
\newblock In \emph{International Conference on Artificial Neural Networks},
  pp.\  697--706. Springer, 2007.

\bibitem[Willems et~al.(1995)Willems, Shtarkov, and
  Tjalkens]{willems1995context}
Willems, Frans~MJ, Shtarkov, Yuri~M, and Tjalkens, Tjalling~J.
\newblock The context-tree weighting method: basic properties.
\newblock \emph{IEEE Transactions on Information Theory}, 41\penalty0
  (3):\penalty0 653--664, 1995.

\bibitem[Wu et~al.(2017)Wu, Mansimov, Grosse, Liao, and Ba]{wu2017scalable}
Wu, Yuhuai, Mansimov, Elman, Grosse, Roger~B, Liao, Shun, and Ba, Jimmy.
\newblock Scalable trust-region method for deep reinforcement learning using
  {K}ronecker-factored approximation.
\newblock In \emph{Advances in neural information processing systems}, pp.\
  5285--5294, 2017.

\bibitem[Zhang et~al.(2015)Zhang, Levine, McCarthy, Finn, and
  Abbeel]{zhang2015policy}
Zhang, Marvin, Levine, Sergey, McCarthy, Zoe, Finn, Chelsea, and Abbeel,
  Pieter.
\newblock Policy learning with continuous memory states for partially observed
  robotic control.
\newblock \emph{CoRR}, 2015.

\bibitem[Zhu et~al.(2017)Zhu, Li, and Poupart]{zhu2017improving}
Zhu, Pengfei, Li, Xin, and Poupart, Pascal.
\newblock On improving deep reinforcement learning for {POMDP}s.
\newblock \emph{arXiv preprint 1704.07978}, 2017.

\end{thebibliography}
